%% file: neurips_2026.tex
\documentclass{article}

\PassOptionsToPackage{numbers, compress}{natbib}

\usepackage[main, final]{neurips_2026}

\usepackage[utf8]{inputenc} %
\usepackage[T1]{fontenc}    %
\usepackage{hyperref}       %
\usepackage{url}            %
\usepackage{booktabs}       %
\usepackage{amsfonts}       %
\usepackage{nicefrac}       %
\usepackage{microtype}      %

\usepackage{mathtools,bm}
\usepackage{amsthm}
\usepackage{graphicx}
\usepackage{wrapfig}
\usepackage[table,xcdraw]{xcolor}
\definecolor{PTALinkPink}{HTML}{FF3399}
\hypersetup{
  colorlinks=true,
  linkcolor=red,
  citecolor=PTALinkPink,
  urlcolor=PTALinkPink,
}
\definecolor{PTAStudent}{HTML}{006C70}
\definecolor{PTASelection}{HTML}{68449E}
\newcommand{\ptaStudentTerm}[1]{\textcolor{PTAStudent}{#1}}
\newcommand{\ptaSelectionTerm}[1]{\textcolor{PTASelection}{#1}}
\usepackage[normalem]{ulem}
\usepackage{algorithm}
\usepackage{algpseudocode}
\usepackage{subcaption}
\usepackage[inline]{enumitem}
\usepackage{multirow}
\useunder{\uline}{\ul}{}
\usepackage{tikz}
\usepackage{siunitx}
\usetikzlibrary{shadows, matrix, positioning, arrows.meta, shapes, calc, fit, backgrounds}

\makeatletter
\def\fps@figure{!t}
\def\fps@table{!t}
\makeatother

\theoremstyle{plain}
\newtheorem{theorem}{Theorem}[section]
\newtheorem{proposition}[theorem]{Proposition}
\newtheorem{corollary}[theorem]{Corollary}
\theoremstyle{definition}
\newtheorem{definition}[theorem]{Definition}

\title{Permute-then-Adapt: Weak-to-Strong \\ Contrastive Image--Text Adaptation}

\author{%
  \begin{tabular}[t]{@{}ccc@{}}
    \makebox[\dimexpr\textwidth/3-2\tabcolsep\relax][c]{Jinhao Li\textsuperscript{1}} &
    \makebox[\dimexpr\textwidth/3-2\tabcolsep\relax][c]{Sarah Erfani\textsuperscript{1}} &
    \makebox[\dimexpr\textwidth/3-2\tabcolsep\relax][c]{Lei Feng\textsuperscript{2}} \\
    {\normalfont\footnotesize\texttt{jinhao.li2@unimelb.edu.au}} &
    {\normalfont\footnotesize\texttt{sarah.erfani@unimelb.edu.au}} &
    {\normalfont\footnotesize\texttt{fenglei@seu.edu.cn}} \\[7pt]
    \makebox[\dimexpr\textwidth/3-2\tabcolsep\relax][c]{Guangrui Li\textsuperscript{3}\thanks{Work done before Guangrui joined AWS.}} &
    \makebox[\dimexpr\textwidth/3-2\tabcolsep\relax][c]{James Bailey\textsuperscript{4}} &
    \makebox[\dimexpr\textwidth/3-2\tabcolsep\relax][c]{Feng Liu\textsuperscript{1}\thanks{Corresponding author.}} \\
    {\normalfont\footnotesize\texttt{guangrli@amazon.com}} &
    {\normalfont\footnotesize\texttt{james.a.bailey@monash.edu}} &
    {\normalfont\footnotesize\texttt{feng.liu1@unimelb.edu.au}} \\[7pt]
    \multicolumn{3}{@{}c@{}}{{\normalfont\small\textsuperscript{1}School of Computing and Information Systems, The University of Melbourne, Australia}} \\
    \multicolumn{3}{@{}c@{}}{{\normalfont\small\textsuperscript{2}School of Computer Science and Engineering, Southeast University, China}} \\
    \multicolumn{3}{@{}c@{}}{{\normalfont\small\textsuperscript{3}AWS AI Labs, USA}} \\
    \multicolumn{3}{@{}c@{}}{{\normalfont\small\textsuperscript{4}Department of Data Science and Artificial Intelligence, Monash University, Australia}}
  \end{tabular}
}

\usepackage{placeins}

\begin{document}

\maketitle

\begin{abstract}
  Image--text alignment models such as CLIP are typically trained with contrastive learning, where unpaired examples are treated as strict negatives.
  While this assumption introduces noise by penalising semantically related pairs, existing solutions often rely on heuristic similarity thresholds that lack statistical grounding and sensitivity to dataset-specific noise distributions.
  In this paper, we propose \emph{Permute-then-Adapt (PTA)}, a framework that leverages efficient \emph{weak models} to identify missed positives while controlling Type-I error relative to a conditional permutation reference, up to calibration approximation error.
  Unlike standard distillation or arbitrary thresholding,
  PTA estimates a random-pairing reference distribution from the weak teacher’s similarity scores and uses its upper tail, determined by $\alpha$, to select auxiliary positives.
  The calibrated positive set is then trained against by a single multi-positive contrastive objective: each anchor maximises the total probability mass it assigns to its positive set.
  Offline calibration with the weak model adds negligible training overhead.
  Experiments show gains over heuristic soft-label approaches in object recognition, cross-modal retrieval, and data efficiency.
  Project repository: \url{https://github.com/JinhaoLee/Permute-then-Adapt}.
\end{abstract}

\input{source/introduction}
\input{source/related_work}
\input{source/prelimiaries}

\input{source/method}

\input{source/exp}

\textbf{Limitations.}
Experiments cover CLIP-based dual encoders, natural-image classification, and image--text retrieval, with scaling up to three million CC3M pairs. Multilingual corpora, specialised domains, and larger pretraining regimes remain to be evaluated. PTA retains observed matches as positives and selects auxiliary positives within each minibatch. Repairing incorrect original pairings or searching broader candidate pools would require extending the reference construction and evaluating the associated computational trade-offs.

\section{Conclusion}
PTA uses a weak teacher to select auxiliary positives for image--text adaptation. Permutation calibration provides Type-I error control relative to the conditional permutation reference, up to calibration approximation error. A frozen dual-encoder teacher enables offline calibration. A single multi-positive objective maximises probability mass over the calibrated positive set and recovers CLIP InfoNCE when auxiliary positives are absent. PTA achieves the highest average classification accuracy among the evaluated weak-supervision methods and improves retrieval over supervised fine-tuning. Under inference-time dropout, student accuracy is less sensitive to moderate corruption than teacher accuracy, while severe corruption still degrades student performance.

\clearpage

\section*{Acknowledgments and Disclosure of Funding}
We thank The University of Melbourne's Research Computing Services and the Petascale Campus for providing computational resources.

\textbf{Funding.} Jinhao Li is supported by the Melbourne Research Scholarship. Feng Liu is supported by the ARC with grant numbers DP230101540, DE240101089 and DP260103996. \par The authors declare no competing interests. \par \bibliographystyle{plainnat}
\bibliography{references}

\appendix

\newpage

\input{source/appendix}

\clearpage

\input{checklist.tex}

\end{document}

%% file: source/introduction.tex
\section{Introduction}

Recent advances in multimodal learning \citep{radford2021learning, jia2021scaling, mu2022slip, xu2023demystifying, fang2023data, zhai2023sigmoid, tschannen2025siglip} have demonstrated the immense potential of image-text alignment models for understanding and connecting visual and textual information. Models like CLIP \citep{radford2021learning} have shown remarkable capabilities in zero-shot recognition \citep{deng2009imagenet}, cross-modal retrieval \citep{lin2014microsoft}, and serving as foundational components for larger multimodal systems \citep{wang2024qwen2}. These models achieve their performance through contrastive learning on large-scale image-text pairs, where the core training objective is to maximise similarity between matched pairs while minimising similarity between unmatched ones.

\begin{figure}[!t]
  \centering
  \includegraphics[width=0.98\linewidth]{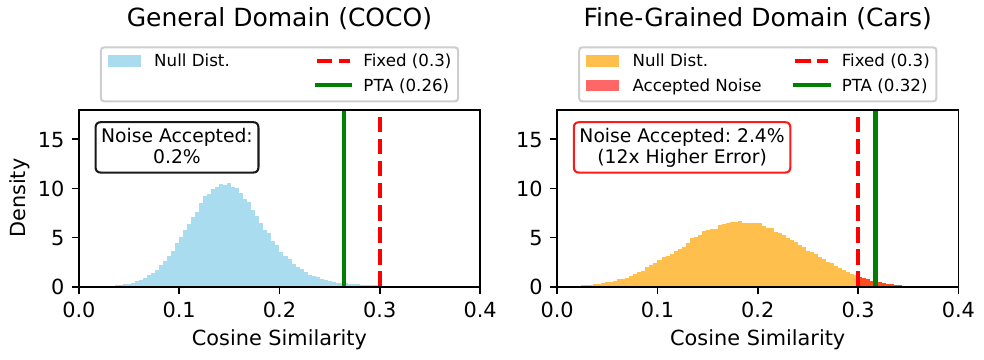}
  \caption{\textbf{The risk of uncalibrated heuristics.}
    While a fixed threshold (red dashed) works on COCO (\textbf{Left}), it fails on fine-grained Stanford Cars (\textbf{Right}), where the random-pairing null distribution shifts rightward.
    The fixed threshold accepts \textbf{12$\times$ more false positives} from the null on Cars.
  In contrast, \textbf{PTA} (green solid) estimates a dataset-specific permutation null and adapts its cutoff to control Type-I error relative to that reference at level $\alpha$, up to calibration approximation error.}
  \label{fig:motivation}
  \vspace{-0.75em}
\end{figure}

However, the assumption that all unmatched pairs are completely unrelated is a fundamental limitation in contrastive learning, as it forces the model to treat semantically valid overlaps as false negatives \citep{byun2024mafa, xing2024vision}.
Prior work uses similarity-based filtering to curate image--text training data \citep{fang2023data}, or learns soft alignment targets through teacher supervision and momentum distillation \citep{huang2024cross, li2021align}. These strategies motivate dataset-adaptive calibration of auxiliary-positive selection.
A fixed threshold is inherently brittle: it fails to adapt to the varying signal-to-noise ratios across different datasets or model architectures, leading to either missed positives or the introduction of noise.
This reveals a critical gap: the lack of a mechanism to distinguish potential signals from the null distribution of random pairings without manual tuning.
As shown in Figure \ref{fig:motivation}, a fixed threshold that works for a general domain like COCO fails catastrophically on fine-grained tasks (e.g., Stanford Cars), accepting 12$\times$ more noise due to the domain's naturally higher baseline similarities.
We address this by proposing a \emph{statistically grounded} framework.
Our method uses permutation testing to estimate the conditional random-pairing null and select auxiliary positives with Type-I error control relative to that reference, up to calibration approximation error.

In this paper, we propose a framework for enhancing image-text alignment by leveraging the semantic knowledge of \emph{efficient weak models}.
Our motivation stems from the insight that weak-to-strong generalisation \citep{burns2023weak} (e.g., smaller CLIP variants), despite their limitations in zero-shot accuracy, encode rich signals about semantic relatedness that are computationally cheap to access.
However, raw similarity scores are noisy; relying on a single fixed threshold to extract them is brittle and fails to generalise across domains.
To rigorously separate this signal from noise, we introduce \textbf{Permute-then-Adapt (PTA)}, a weak-to-strong framework that replaces heuristic thresholding with \emph{statistical hypothesis testing}.
Inspired by the permutation test \citep{hemerik2018exact},
PTA estimates a dataset-specific reference distribution of weak-teacher similarities under random image–text pairings and selects auxiliary positives whose scores exceed its calibrated upper-tail threshold.
This approach requires no additional validation data and adapts naturally to the intrinsic noise floor of different datasets.
The resulting calibrated positive set is then trained against by a single multi-positive contrastive objective: for each anchor we maximise the total probability mass assigned to its positive set, where the positive set is the ground-truth diagonal augmented with the calibrated off-diagonal pairs.
We validate our approach through extensive experiments across image classification and retrieval benchmarks, demonstrating that our statistically grounded guidance consistently outperforms heuristic baselines.

Our main contributions can be summarised as follows:
\begin{enumerate*}[label=(\roman*)]
  \itemsep0em
\item \emph{Permutation-calibrated auxiliary positive selection.} We use semantically irrelevant permutations to calibrate a dataset-adaptive cutoff, providing Type-I error control relative to the conditional random-pairing reference at significance level $\alpha$, up to Monte Carlo and percentile approximation error.
\item \emph{PTA multi-positive objective.} We train against the calibrated set with a single positive-set marginal-likelihood loss.
\item Through extensive experiments on multiple zero-shot image classification benchmarks and two cross-modal retrieval datasets, we show that PTA consistently outperforms existing weakly supervised baselines and tolerates moderate inference-time teacher dropout.
\end{enumerate*}

%% file: source/related_work.tex
\section{Related work}\label{sec:relatedwork}

\textbf{Vision-language models.}
Since the introduction of CLIP \citep{radford2021learning}, numerous models building upon its multimodal representation learning framework have emerged. Subsequent work has explored architectural improvements and training-recipe refinements, such as ALIGN \citep{jia2021scaling} which scales up the training data and model size, EVA-CLIP \citep{sun2023eva} which improves the training stability and efficiency of large-scale CLIP through optimised initialisations and learning rates, and Florence \citep{yuan2021florence} which incorporates additional visual reasoning capabilities. BLIP \citep{li2022blip} uses caption bootstrapping for unified vision--language understanding and generation, while LiT \citep{zhai2022lit} trains a text encoder against a locked pretrained image encoder. FG-CLIP \citep{xie2025fg} explicitly targets fine-grained visual--textual alignment. PTA is orthogonal to this line of work: rather than scaling the model or replacing the backbone, it provides a post-hoc refinement that exploits a calibrated, dataset-adaptive selection of auxiliary positives and could in principle be applied on top of any of these architectures.

\textbf{CLIP fine-tuning.} Recent advancements in fine-tuning CLIP models have led to various parameter-efficient methods aimed at enhancing performance on downstream tasks. The Context Optimization (CoOp) \citep{zhou2022learning} approach introduces learnable text prompts, enabling the model to adapt to specific tasks by optimizing these prompts during training. Building upon this, Conditional Prompt Learning (CoCoOp) \citep{zhou2022conditional} extends CoOp by conditioning the prompts on image instances, thereby improving generalisation across diverse visual concepts. Alternatively, CLIP-Adapter \citep{gao2024clip} incorporates lightweight residual-style adapters into CLIP's architecture, facilitating the learning of task-specific features without extensive retraining. Tip-Adapter \citep{zhang2021tip} further streamlines this process by constructing a key-value cache model from the few-shot training set, enabling efficient adaptation with minimal or no training. Additionally, methods like Visual Prompt Tuning (VPT) \citep{jia2022visual} add learnable prompt tokens in the input space, fine-tuning the model effectively for various tasks. These strategies aim to improve CLIP's adaptability and performance while maintaining computational efficiency.

\textbf{Bootstrapping and pseudo-labelling.}
Several works attempt to mitigate noise in image-text alignment through bootstrapping or pseudo-labelling. BLIP \citep{li2022blip} employs caption bootstrapping and multi-stage training to refine noisy web-crawled supervision. More recent studies propose pseudo-labelling strategies to relax one-hot supervision, such as Cross-Modal Similarity Regulation \citep{jiang2023vision}, multi-label pseudo-labelling \citep{xing2024vision}, and soft-label alignment \citep{huang2024cross}. These methods highlight the importance of leveraging softer or multiple alignment signals. Our approach shares this motivation but differs by introducing a statistically grounded mechanism (permutation testing) to identify semantically meaningful extra positives with controlled false alarm rates.

\textbf{Knowledge distillation and weak-to-strong supervision.}
Knowledge distillation techniques, such as ALBEF \citep{li2021align}, employ momentum encoders to refine alignment signals from auxiliary models. Weak-to-strong transfer has also been studied more broadly \citep{burns2023weak}, emphasising how weaker models guide stronger ones. For CLIP classification, CPL~\citep{li2025exploring} learns class prototypes under weak-model supervision. PTA builds upon this weak-to-strong philosophy but departs from direct distillation: instead of transferring continuous similarity scores, it adaptively thresholds weak-model signals using a permutation-derived null distribution, producing statistically validated supervision.

\textbf{Mitigating false negatives.}
False negatives, i.e.\ unpaired image–text instances that are semantically related, have recently received direct attention in vision–language pretraining. For instance, MAFA \citep{byun2024mafa} proposes an efficient connection mining process to convert some hard false negatives into positives, and smooths the contrastive ITC loss to reduce over-penalisation. Empirically, MAFA shows that doing this yields substantial downstream gains, especially under hard-negative sampling strategies. PTA distinguishes itself by employing a permutation-based null distribution to statistically validate and promote high-scoring unpaired tuples as auxiliary positives for weak-to-strong supervision, and by reusing the same calibrated mask to clean false negatives from the contrastive loss.
Related approaches mitigate noisy supervision by modifying contrastive objectives or learning softer alignment targets. PyramidCLIP~\citep{gao2022pyramidclip}, SoftCLIP~\citep{gao2024softclip}, and CLIP-PSD~\citep{andonian2022robust} follow this direction. PTA calibrates auxiliary-positive selection against a permutation reference.
AdaCL~\citep{pan2025discovering} focuses on \emph{clone negatives} captions that are semantically close but not exact matches by adaptively reshaping the positive logit, but still treats all non-matching pairs as negatives.
NCU~\citep{han2025unlearning} studies post-hoc unlearning of noisy correspondences in pretrained CLIP, addressing both false positives and false negatives, and can operate with a subset of the original pretraining data.
We instead study post-hoc weak-to-strong refinement on a small, clean downstream dataset with correct positives, where in-batch sampling introduces hard false negatives into the contrastive loss.

%% file: source/prelimiaries.tex
\section{Preliminaries}
\label{sec:pre}

\textbf{Dual encoder models.} Dual encoder models \citep{radford2021learning} belong to a class of neural architectures designed to learn representations of two different modalities in a shared embedding space. These models consist of two separate encoders that process different input types (e.g., images and text) and project them into a common semantic space where similarity can be directly computed. Formally, we define: an image encoder $\ptaStudentTerm{f_\text{image}}: \mathcal{X} \rightarrow \mathbb{R}^d$ that maps images to a $d$-dimensional embedding, and a text encoder $\ptaStudentTerm{f_\text{text}}: \mathcal{Z} \rightarrow \mathbb{R}^d$ that maps text to the same $d$-dimensional space.

\textbf{Image-text alignment.} Image-text alignment aims to capture semantic correspondences between visual and textual content~\citep{radford2021learning,li2024visual}. Given a dataset $\mathcal{D} = \{(I_i, T_i)\}_{i=1}^N$ of image-text pairs, where $I_i \in \mathcal{X}$ represents an image and $T_i \in \mathcal{Z}$ represents its corresponding text description, the goal is to learn a mapping that reflects semantic similarity in a shared embedding space. For notational brevity, we denote the similarity score assigned by a model $\ptaStudentTerm{f}$ to an image-text pair $(I_i, T_j)$ as the inner product of their normalised embeddings: $\ptaStudentTerm{f(I_i, T_j)} = \langle \ptaStudentTerm{f_\text{image}(I_i)}, \ptaStudentTerm{f_\text{text}(T_j)} \rangle$.

We formulate the alignment task using contrastive learning on the full dataset $\mathcal{D}$. The objective aligns embeddings by maximising the similarity score $\ptaStudentTerm{f(I_i, T_i)}$ for matched pairs while minimising it for mismatched pairs (e.g., $I_i$ and $T_j$ where $i \neq j$).

To ensure proper gradient scaling, a learnable temperature parameter $\ptaStudentTerm{\tau}$ is introduced. For each anchor $i$, the image-to-text and text-to-image contrastive losses are defined as:
\begin{equation}
  \ptaStudentTerm{\ell_{\rm i2t}^{(i)}} = -\log \frac{\ptaStudentTerm{\exp\left( f(I_i, T_i) / \tau \right)}}{\sum_{j=1}^N \ptaStudentTerm{\exp\left( f(I_i, T_j) / \tau \right)}},
  \ptaStudentTerm{\ell_{\rm t2i}^{(i)}} = -\log \frac{\ptaStudentTerm{\exp\left( f(I_i, T_i) / \tau \right)}}{\sum_{j=1}^N \ptaStudentTerm{\exp\left( f(I_j, T_i) / \tau \right)}}.
\end{equation}
The total loss
\begin{align}
  \label{eq:totalloss_paired}
  \ptaStudentTerm{\mathcal{L}^{\rm clip}} = \frac{1}{2N} \sum_{i=1}^N \left( \ptaStudentTerm{\ell_{\rm i2t}^{(i)}} + \ptaStudentTerm{\ell_{\rm t2i}^{(i)}} \right)
\end{align}
is minimised over $\mathcal{D}$ to align embeddings, with negative pairs enforcing discriminative separation.

%% file: source/method.tex
\newcommand{\ptaMatrixFont}[1]{\bm{\mathsf{#1}}}

\begin{figure}[!t]
  \centering
  \includegraphics[width=1\linewidth]{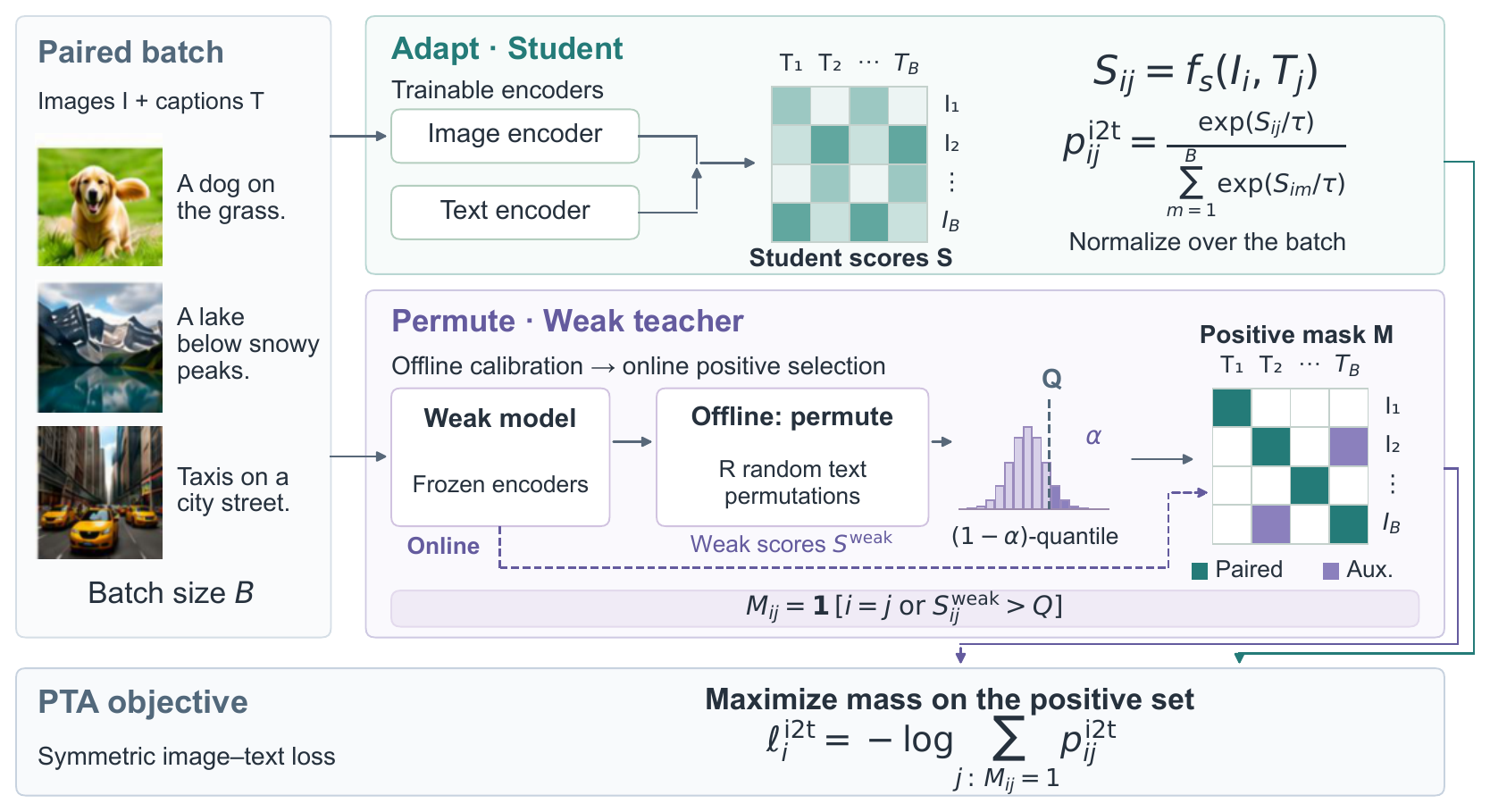}
  \caption{
    \textbf{Overview of Permute-then-Adapt (PTA).}
    Given paired image--text samples, PTA uses a frozen weak teacher to estimate a permutation null
    distribution and dataset-adaptive threshold. This threshold selects calibrated off-diagonal
    auxiliary positives, which join ground-truth diagonal pairs to form positive sets.
    The student is optimized with a positive-set marginal likelihood objective.
  }
  \label{fig:pta_framework}
\end{figure}

\section{Image-text alignment with weak models}

We introduce our method to improve a strong model's image-text alignment using weak models.

\subsection{Our method: Permute-then-Adapt}\label{sec:pta}
In this section, we will discuss the problem setup and our method. \figureautorefname~\ref{fig:pta_framework} demonstrates our method visually.
The overall algorithm can be found in Algorithm~\ref{alg:pta}.

A key limitation of standard contrastive learning (\equationautorefname~\ref{eq:totalloss_paired}) is its hard-negative assumption: all non-paired tuples $(I_i,T_j)$ with $i\neq j$ are treated as negatives, even though some may be semantically related.
A natural idea is to threshold teacher $\ptaSelectionTerm{f}$ similarity scores to convert such tuples into auxiliary positives;
however, fixed cutoffs \citep{fang2023data} are brittle across datasets and model architectures.

\textbf{Problem setup. }
We consider post-hoc refinement of a pretrained dual-encoder (the \emph{student} $\ptaStudentTerm{f^\text{s}}$, also denoted the \emph{strong model}) on a paired dataset $\mathcal{D}$.
Given a frozen, lightweight \emph{teacher} $\ptaSelectionTerm{f}$ (also denoted the \emph{weak model}\footnote{We use ``teacher''/``student'' and ``weak''/``strong'' interchangeably throughout the paper. The framework only requires that the teacher is a frozen dual-encoder whose similarity scores can serve as the calibration signal; nothing in our derivation relies on it being smaller than the student. We choose a smaller teacher in our experiments for computational practicality.}),
our goal is to identify a subset of non-paired tuples $(I_i,T_j)$ with $i\neq j$ as \emph{auxiliary positives}
and use them to enhance the image-text alignment of the student $\ptaStudentTerm{f_\text{s}}$.

\textbf{Weak teacher similarities}.
A data-calibrated threshold $\ptaSelectionTerm{Q}$ defines pseudo-labels for non-paired tuples.
Standard contrastive learning encodes supervision via the dataset pairing indicator
\begin{equation}
  l_D(I_i,T_j) := \mathbb{I}[i=j],
\end{equation}
Unpaired tuples receive the hard label $l_D=0$, which we replace with a weak pseudo-label $\ptaSelectionTerm{l_f(\cdot,\cdot;Q)}$:
\begin{align}
  \label{eq:weakmodel}
  \ptaSelectionTerm{l_{f}(I,T;Q)}=
  \begin{cases}
    1 & \text{if } \ptaSelectionTerm{f(I,T)} > \ptaSelectionTerm{Q}, \\
    0 & {\rm otherwise},
  \end{cases}
\end{align}
where $\ptaSelectionTerm{f}$ is a frozen weak teacher that scores arbitrary image--text pairs, and $\ptaSelectionTerm{Q}$ selects high-scoring tuples as auxiliary positives.
Assigning the weak pseudo-label $\ptaSelectionTerm{l_f(I,T;Q)}$ gives the adaptation set
\begin{align}
  \label{eq:trainset_weak}
  \mathcal{D}_{\rm adapt}^{\ptaSelectionTerm{f}}(\ptaSelectionTerm{Q})=\{(I_i,T_j,\ptaSelectionTerm{l_{f}(I_i,T_j;Q)})\}_{i,j=1}^N.
\end{align}
We can then fine-tune the student using $\mathcal{D}_{\rm adapt}^{\ptaSelectionTerm{f}}(\ptaSelectionTerm{Q})$\footnote{Although Eq.~\ref{eq:trainset_weak} defines the full candidate set of $N^2$ image–text pairs, in practice we do not materialise all possible pairs during training. Instead, training operates on mini-batches}.
A key difficulty is that $\mathcal{D}_{\rm adapt}^{\ptaSelectionTerm{f}}(\ptaSelectionTerm{Q})$ is highly sensitive to the choice of $\ptaSelectionTerm{Q}$.
Fixed similarity thresholds \citep{fang2023data} are not comparable across datasets or model architectures: the same numeric value of $\ptaSelectionTerm{Q}$
may correspond to very different signal-to-noise ratios, leading to either missed auxiliary positives or excessive noise (as shown in \figureautorefname~\ref{fig:motivation}).
To make selection principled and dataset-adaptive, we calibrate $\ptaSelectionTerm{Q}$ via permutation testing at significance level $\alpha$, providing Type-I error control relative to the conditional permutation reference, up to Monte Carlo and percentile approximation error \citep{hemerik2018exact}.

\subsubsection{Permute: estimating the null distribution}

\textbf{Permutation-based null.}
Selecting $\ptaSelectionTerm{Q}$ requires a reference distribution of similarity scores for \emph{irrelevant} image--text pairs.
Under the independence null $H_0: I \perp T$, text permutation preserves the observed marginals but breaks associations,
yielding a valid (conditional) permutation null \citep{gretton2007kernel, hemerik2018exact}.

\textbf{Interpretation.}
The permutation null is intentionally \emph{not} a generative model of hard semantic negatives, and our framework does not require that real images and texts are independent. Instead, the null serves as an \emph{empirical reference distribution of weak-similarity scores under random mismatching}, evaluated on the same dataset and the same weak teacher used downstream. PTA then promotes off-diagonal pairs whose weak similarity is implausibly large relative to this reference; semantically related hard negatives, which would otherwise be penalised by the standard contrastive loss, are precisely the right-tail deviations we wish to rescue from strict negation. The null thus provides a conservative, dataset-adaptive baseline without assuming that semantic clustering is absent.

\textbf{Semantically irrelevant perturbations.}
Shuffling text in the paired dataset $\mathcal{D}=\{(I_i,T_i)\}_{i=1}^N$ yields an empirical permutation null \emph{sample} of weak similarity scores.
Let $\mathfrak{G}_N$ denote the permutation group on $[N]=\{1,\dots,N\}$, with elements $\sigma:[N]\to[N]$ acting on the text tuple $\mathcal{T}=(T_1,\dots,T_N)$ via
$\sigma\mathcal{T} := (T_{\sigma(1)},\dots,T_{\sigma(N)})$.
Under semantically irrelevant image--text permutations, we draw $R$ random permutations
$\{\sigma_r\}_{r=1}^R \sim \mathrm{Unif}(\mathfrak{G}_N)$ and collect the resulting null similarity scores:
\begin{align}
  \label{eq:null}
  \ptaSelectionTerm{\mathcal{S}_{\mathrm{perm}}}
  ~:=~
  \big\{ \ptaSelectionTerm{f(I_i, T_{\sigma_r(i)})} ~:~ i \in [N],~ r \in [R] \big\}.
\end{align}
This empirical permutation null sample retains repeated scores; we threshold auxiliary positives with Type-I error control relative to that reference, up to calibration approximation error.

\begin{proposition}[Validity of the permutation null \citep{good2005permutation, lehmann2005testing, hemerik2018exact}]
  \label{prop:null_validity}
  Under the independence null $H_0:I\perp T$, suppose text assignments are exchangeable conditional on the observed image and text marginals $\mathcal M$.
  All matchings $\{(I_i,T_{\sigma(i)})\}_{i=1}^{N}$ are then equally likely under $H_0$.
  For a fixed teacher $\ptaSelectionTerm{f}$, a uniformly sampled matching and image index yield the conditional random-pairing score CDF
  \[
    F_0(t)=\frac{1}{N^2}\sum_{i,j=1}^{N}
      \mathbf{1}\{\ptaSelectionTerm{f(I_i,T_j)}\le t\}.
  \]
  Independent uniform text permutations provide a Monte Carlo approximation to this reference distribution.
  When calibration uses a subset, apply the statements below with $N=N_{\mathrm{cal}}$ and $\mathcal M$ denoting its observed marginals.
\end{proposition}

Conditional permutation testing provides the theoretical background \citep{lehmann2005testing}; the bound below explicitly accounts for calibration approximation error.

\begin{definition}[Permutation-Calibrated Selection]
  \label{def:pta_selection}
  Given a significance level $\alpha\in(0,1)$, define
  \[
    \ptaSelectionTerm{Q_{\mathrm{PTA}}}
    :=\mathrm{percentile}_{1-\alpha}
      (\ptaSelectionTerm{\mathcal{S}_{\mathrm{perm}}}).
  \]
  The selection rule assigns a pseudo-label according to
  \begin{equation}
    \ptaSelectionTerm{l^{p}_{f}(I,T)}\coloneqq
    \begin{cases}
      1 & \text{if }\ptaSelectionTerm{f(I,T)}>\ptaSelectionTerm{Q_{\mathrm{PTA}}},\\
      0 & \text{otherwise}.
    \end{cases}
  \end{equation}
\end{definition}

\begin{corollary}[Type-I error control]
  \label{cor:error_control}
  Condition on $\mathcal M$, the fixed teacher $\ptaSelectionTerm{f}$, and the sampled permutations $\Pi$.
  Let $J,K$ be independent uniform indices in $[N]$, independent of $\Pi$, so that $(I_J,T_K)$ is a random pair from the same observed marginals.
  Write $Q=Q_{\mathrm{PTA}}$, and let $\widehat F_R$ be the empirical CDF of $\mathcal S_{\mathrm{perm}}$.
  Define the finite-calibration approximation error
  \[
    \varepsilon_{\mathrm{cal}}
    :=\|\widehat F_R-F_0\|_\infty
      +[1-\widehat F_R(Q)-\alpha]_+,
  \]
  where $[x]_+=\max\{x,0\}$.
  Then the Type-I error relative to the conditional random-pairing reference satisfies
  \begin{equation}
    \Pr_{J,K}\!\left(
      \ptaSelectionTerm{l_f^p(I_J,T_K)}=1
      \mid\mathcal M,\ptaSelectionTerm{f},\Pi
    \right)
    \le\min\{1,\alpha+\varepsilon_{\mathrm{cal}}\}.
    \label{eq:pta_reference_control}
  \end{equation}
  Indeed, the left-hand side equals $1-F_0(Q)$, which is at most
  $1-\widehat F_R(Q)+\|\widehat F_R-F_0\|_\infty$.
  The first error term is the Monte Carlo reference approximation; the second accounts for percentile discretisation or interpolation and is zero for a generalised-inverse empirical quantile.
\end{corollary}

Thus, $\alpha$ specifies the tolerated Type-I error relative to the conditional permutation reference, up to the stated calibration approximation error.
This reference control does not assert semantic precision of the selected auxiliary positives.

\subsubsection{Adapt: enhance image-text alignment}
\label{sec:Adapt}

Following this selection procedure, we collect the permutation-calibrated pseudo-labels into a training set of additional positives for fine-tuning the strong model.
Let $\mathcal{K}\subset [N]$ be a set of indices with $|\mathcal{K}|=B$ defining a mini-batch.
We use paired samples from $\mathcal{D}$ together with off-diagonal PTA-labelled pairs mined within the batch:
\begin{equation}
  \mathcal{D}_{\text{batch}} := \{(I_k, T_k): k \in \mathcal{K}\},
  \mathcal{D}^{\text{PTA}}_{\text{batch}} := \{(I_k, T_m, \ptaSelectionTerm{l_f^p(I_k,T_m)}) : k,m \in \mathcal{K},~ k\neq m\}.
\end{equation}
By construction, $\mathcal{D}^{\text{PTA}}_{\text{batch}} \subset \mathcal{D}^{\ptaSelectionTerm{f}}_{\text{adapt}}(\ptaSelectionTerm{Q})$.
Define the PTA indicator for an in-batch pair as
\begin{equation}
  \ptaSelectionTerm{l^{\text{PTA}}_{km}} := \ptaSelectionTerm{l^{p}_{f}(I_k, T_m)}\in\{0,1\}, \qquad (k\neq m),
\end{equation}
and set $\ptaSelectionTerm{l^{\text{PTA}}_{kk}}:=0$ for completeness.
We then form the final target mask
\begin{equation}
  \tilde{l}^{\mathrm{PTA}}_{km} \;=\; \mathbb{I}[k=m] \;\vee\; \ptaSelectionTerm{l^{\mathrm{PTA}}_{km}}, \quad \tilde{l}^{\mathrm{PTA}}_{km} \in\{0,1\}.
\end{equation}
That is, each anchor always includes its ground-truth match ($k=m$) and includes additional in-batch positives selected by PTA via the permutation-calibrated threshold.
In contrast to standard CLIP supervision, which treats only the diagonal pair as positive, PTA promotes a selected subset of off-diagonal in-batch pairs to positives.

\textbf{Positive-set marginal likelihood objective.}
PTA trains the student with a \emph{positive-set marginal likelihood} \citep{feng2020provably, zhang2024candidate}: each anchor maximises the total probability mass it assigns to its calibrated positive set. For an image anchor $I_k$ with positive set
\begin{equation}
  \ptaSelectionTerm{P_k} \;=\; \{k\}\,\cup\,\{\, m\neq k\,:\,\ptaSelectionTerm{l^{\mathrm{PTA}}_{km}}=1\,\},
  \label{eq:positive_set}
\end{equation}
the image-to-text loss is
\begin{equation}
  \ptaStudentTerm{\ell^{(k)}_{\mathrm{i2t}}}
  \;=\; -\log \frac{\sum_{m\in \ptaSelectionTerm{P_k}}\ptaStudentTerm{\exp(f^{\mathrm{s}}(I_k,T_m)/\tau)}}{\sum_{m'=1}^{B}\ptaStudentTerm{\exp(f^{\mathrm{s}}(I_k,T_{m'})/\tau)}},
  \label{eq:total_loss}
\end{equation}
with the symmetric text-to-image counterpart $\ptaStudentTerm{\ell^{(m)}_{\mathrm{t2i}}}$. Averaging both directions over the batch gives
\begin{equation}
  \ptaStudentTerm{\mathcal{L}^{\mathrm{PTA}}}
  \;=\; \frac{1}{2}\!\left(\frac{1}{B}\sum_{k=1}^B \ptaStudentTerm{\ell^{(k)}_{\mathrm{i2t}}} \;+\; \frac{1}{B}\sum_{m=1}^B \ptaStudentTerm{\ell^{(m)}_{\mathrm{t2i}}}\right).
\end{equation}
When $\ptaSelectionTerm{P_k} = \{k\}$ this objective reduces to the standard CLIP InfoNCE loss; the only hyperparameter is the significance level $\alpha$ controlling which off-diagonal pairs join $\ptaSelectionTerm{P_k}$.

\input{source/algo}

%% file: source/algo.tex
\begin{algorithm}[t]
\caption{Permute-then-Adapt (PTA) Training Framework}
\label{alg:pta}
\begin{algorithmic}[1]
\State \textbf{Input:} Teacher $\ptaSelectionTerm{f}$, Student $\ptaStudentTerm{f^s}$, Dataset $\mathcal{D}$, significance level $\alpha$.
\State \textbf{Stage 1: Calibration (Offline)}
\State Sample $N_{\mathrm{cal}}$ image-text pairs from $\mathcal{D}$ and apply $R$ random text permutations to obtain the empirical null sample $\ptaSelectionTerm{\mathcal{S}_{\mathrm{perm}}}$.
\State Determine threshold $\ptaSelectionTerm{Q_{\mathrm{PTA}}} \leftarrow \text{percentile}_{1-\alpha}(\ptaSelectionTerm{\mathcal{S}_{\mathrm{perm}}})$.
\State \textbf{Stage 2: Adaptation (Online)}
\For{mini-batch $\mathcal{B}=\{(I_k,T_k)\}_{k=1}^B$}
    \State \textcolor{gray}{// Calibrated positive set}
    \State Compute teacher similarity matrix $\ptaSelectionTerm{\ptaMatrixFont{S}_{\mathrm{weak}}}\in\mathbb{R}^{B\times B}$.
    \State For each anchor $k$, form
    $\ptaSelectionTerm{P_k}\leftarrow\{k\}\cup\{m\!\neq\! k : \ptaSelectionTerm{S_{\mathrm{weak}}(I_k,T_m)}>\ptaSelectionTerm{Q_{\mathrm{PTA}}}\}$ (Eq.~\ref{eq:positive_set}).
    \State \textcolor{gray}{// Single-equation multi-positive contrastive loss}
    \State Compute $\ptaStudentTerm{\mathcal{L}^{\mathrm{PTA}}}$: for each anchor, the negative log of the student's total probability mass on $\ptaSelectionTerm{P_k}$ (Eq.~\ref{eq:total_loss}); standard CLIP InfoNCE is recovered when $\ptaSelectionTerm{P_k}=\{k\}$.
    \State \textbf{Update Student:} $\ptaStudentTerm{\theta} \leftarrow \ptaStudentTerm{\theta} - \eta\, \ptaStudentTerm{\nabla_{\theta}\mathcal{L}^{\mathrm{PTA}}}$.
\EndFor
\end{algorithmic}
\end{algorithm}

%% file: source/exp.tex
\section{Experiments}\label{sec:exp}
We evaluate PTA on zero-shot classification and Karpathy-split cross-modal retrieval,
compare it with heuristic weak supervision, and examine how calibrated
positives enter the objective. Table~\ref{tab:cc3m-cls} reports five-seed
mean$\pm$SD; additional ablations examine teacher quality and calibration.

\input{tables/classification_results} \textbf{Datasets.}
Our experiments leverage the Microsoft COCO \citep{lin2014microsoft} (Common Objects in Context) 2014 dataset.
To evaluate generalisation, we test on natural-image object-recognition tasks spanning CIFAR-10 \citep{krizhevsky2009learning}, CIFAR-100 \citep{krizhevsky2009learning}, Food-101 \citep{bossard2014food}, ImageNet \citep{deng2009imagenet} together with its ImageNet-V2/A/R variants~\citep{recht2019imagenet, hendrycks2021nae, hendrycks2021many}, Describable Textures (DTD) \citep{cimpoi2014describing}, StanfordCars~\citep{Krause_2013_ICCV_Workshops}, CUB-200~\citep{wah2011caltech}, FGVC Aircraft~\citep{maji13fine-grained}, Flowers-102~\citep{Nilsback08}, and Oxford Pets~\citep{parkhi12a}, which cover diverse domains. For cross-modal retrieval, we use Flickr30K \citep{plummer2015flickr30k} and MS-COCO \citep{lin2014microsoft}, standard benchmarks for image-text matching.

\textbf{Baselines.}
We compare our method against a comprehensive set of baselines:
(1) CLIP supervised fine-tuning (SFT) baselines. These provide reference points for direct adaptation of student.
(2) Weak-to-Strong Augmentation (WSA). Proposed by ~\citet{burns2023weak}, WSA incrementally augments positives discovered by weaker models in multi-stage training. This is the closest prior to our weak-to-strong framework.
(3) Soft Align (SA). Following recent work on soft-label alignment~\citep{huang2024cross}, we use similarity scores from a weaker model to generate soft positive labels, replacing the binary supervision in the contrastive loss.
(4) VL PseudoLabel (VPL). Inspired by multi-label pseudo-labeling approaches for vision-language models~\citep{xing2024vision}, this baseline allows each image to have multiple captions marked positive if predicted above a confidence threshold.
(5) Fixed Threshold (FT). As a naive baseline, we apply a constant threshold (e.g., 0.3) on weak model similarity scores to decide additional positives \citep{fang2023data}. This tests whether our permutation-based statistical control is necessary.
(6) Momentum Distillation (MD). Following ALBEF~\citep{li2021align}, we maintain a momentum-updated copy of the weak model as a teacher, and minimize KL divergence between the teacher’s and student’s similarity distributions.

\textbf{Implementation details.}
Unless otherwise specified, a frozen CLIP ViT-B/32 teacher guides a CLIP ViT-B/16 student on COCO train2014. We train for up to 20 epochs using AdamW (learning rate $10^{-4}B/4096$, weight decay $10^{-4}$, 10\% warmup, cosine decay), BF16 mixed precision, and gradient clipping on one GPU (Appendix~\ref{app:compute_resources}). Legacy single-seed runs tune the last four vision blocks. The Table~\ref{tab:cc3m-cls} runs use the full vision tower and last two text blocks, $B=512$, seeds $\{41,42,43,44,45\}$, and the checkpoint with the best in-domain validation recall.
For PTA in Table~\ref{tab:cc3m-cls}, calibration uses 5,000 records in batches of 1,000 at $\alpha=0.01$, discarding fixed-point scores and retaining repetitions. Given $M$ retained scores, we use the $k$-th order statistic with $k=\lceil(M+1)(1-\alpha)\rceil$ and strict selection $s>Q$. This finite-rank variant retains the positive-set objective in Eq.~\ref{eq:total_loss}. Other experiments retain the original empirical 99th-percentile calibration unless stated otherwise.

\input{tables/retrieval_results}

\textbf{Main results.}
Table~\ref{tab:cc3m-cls} reports five-seed zero-shot classification on 13 benchmarks after fine-tuning on COCO. PTA achieves the highest average accuracy, $62.59\%$, and the best mean on 10 of the 13 tasks. SA is strongest on DTD and CUB-200, while WSA is strongest on ImageNet-A.

PTA exceeds SA, the strongest baseline by average accuracy, by $0.80$ percentage points ($62.59\%$ versus $61.80\%$). Relative to SA, PTA gains $3.19$ points on ImageNet-A and $2.35$ on FGVC Aircraft, while trailing by $0.04$ on DTD and $0.72$ on CUB-200. These differences are computed from unrounded five-seed means.

Table~\ref{tab:cc3m-ret} reports cross-modal retrieval on Flickr30k and
MS-COCO. On MS-COCO, PTA raises image-to-text recall@1 from $22.69\%$ to
$25.24\%$ and text-to-image recall@1 from $21.94\%$ to $25.06\%$,
gains of $2.55$ and $3.12$ percentage points over SFT. Recall@5 gains are
$4.24$ and $4.44$ points. On Flickr30k, recall@1 gains are $1.40$ points
for image-to-text and $3.50$ for text-to-image. PTA achieves the highest
text-to-image recall@1, while WSA remains stronger on
image-to-text recall@1, showing that gains vary with retrieval direction.

\begin{wrapfigure}[21]{r}{0.53\textwidth}
  \vspace{-6pt}
  \centering
  \includegraphics[width=\linewidth]{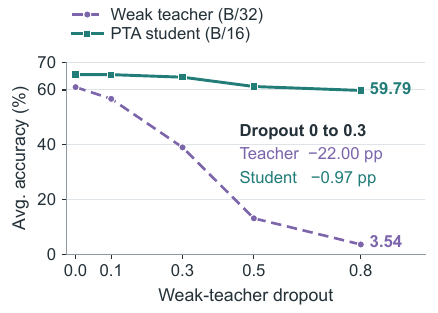}
  \caption{\textbf{Sensitivity to weak-teacher corruption.} Teacher and PTA student accuracy under inference-time teacher dropout, with the student setup fixed. Points reproduce reported benchmark averages; per-task results are in Appendix~\ref{app:weak_teacher_quality}.}
  \label{fig:weak-dropout}
  \vspace{-6pt}
\end{wrapfigure}

\textbf{How weak can the teacher be?}
We progressively corrupt the CLIP-B/32 teacher at inference time with
dropout rates $\{0,0.1,0.3,0.5,0.8\}$, keeping the CLIP-B/16 student
architecture and training hyperparameters fixed. Figure~\ref{fig:weak-dropout}
  compares the teacher's standalone accuracy with the PTA-trained student's
benchmark average. At dropout $0.3$, teacher accuracy falls from $61.00\%$
to $39.00\%$, whereas student accuracy falls from $65.60\%$ to $64.63\%$
($0.97$ percentage points). Student accuracy is less sensitive to this
moderate corruption than the teacher's standalone accuracy. Severe corruption
still matters: at dropout $0.8$, teacher accuracy is $3.54\%$ and student
accuracy falls to $59.79\%$, $5.81$ points below its uncorrupted-teacher
counterpart. Thus, the student tolerates moderate degradation but remains
dependent on useful teacher information. Inference-time dropout is a
controlled corruption proxy; these results do not establish robustness to
every source of teacher error. Full per-task results and a separate
calibration-only diagnostic are in Appendix~\ref{app:weak_teacher_quality}.

\input{tables/mechanism_ablation}

\textbf{Mechanism decomposition.}
Table~\ref{tab:mech-ablation} reports independent ablation runs with matched SFT controls, varying how the calibrated mask is consumed: (i) a \emph{CN-only} variant that uses the mask only to remove false negatives from the contrastive denominator, and (ii) an \emph{asymmetric} variant that keeps the full multi-positive numerator only on the image$\to$text direction. CN-only accounts for roughly 71\% of the full PTA gain ($+1.08\pm 0.06$); admitting the calibrated pairs into the numerator (full PTA) buys the remaining $\sim$0.45.

CN-only retains the observed match as the sole numerator term while
removing selected off-diagonal pairs from the denominator. Full PTA
retains the full normalization and maximizes probability mass over the
calibrated positive set, without requiring equal probability for each
candidate. The asymmetric result is consistent with a benefit from
applying the positive-set objective in both directions.

\textbf{Empirical calibration across datasets and teachers.}
Across 100 resampling seeds on COCO and Flickr30k with CLIP ViT-B/32 and
ViT-B/16 teachers, held-out random-pairing exceedance at $\alpha=0.01$
is $1.005$--$1.023\%$ using $M=5{,}000$ calibration scores. Larger budgets
reduce resampling variability. These global-permutation diagnostics are
separate from training-time minibatch calibration; dependent pooled scores
and finite cached pools limit their interpretation. These empirical rates do not establish exact $\alpha$ control or semantic precision. Full protocols and
results are in Appendix~\ref{app:heldout_random_pairing}; additional
permutation-stability results are in Appendix~\ref{app:global_permutation_stability}.

\newpage

\noindent\begin{minipage}[t]{0.47\textwidth}
\vspace{0pt}
\setlength{\parskip}{5.5pt}
\textbf{Visual grounding.}
Figure~\ref{fg:gradcam_examples} compares Grad-CAM~\citep{selvaraju2016grad}
maps for SA and PTA using the same three queries. PTA can reduce conflicting supervision on potentially related pairs through calibrated auxiliary positives, encouraging the student to focus on query-relevant visual regions. Full queries and retrieval rankings are in Appendix~\ref{app:visualisation}.

\textbf{Training efficiency.}
Figure~\ref{fig:training-throughput} compares per-step throughput relative to SFT. Under fixed teacher preprocessing, PTA-Cache reuses frozen embeddings and scores candidates by matrix multiplication, avoiding repeated teacher encoding. It retains the thresholding and positive-set training rules while achieving $95$--$99\%$ of SFT throughput. The trade-off is preprocessing and embedding storage; cache construction is excluded. Details are in Appendix~\ref{app:cache_efficiency}.

\textbf{Sensitivity to $\alpha$.}
The highest reported ImageNet accuracy occurs at $\alpha=0.01$ (Figure~\ref{fig:alpha-sensitivity}).
\end{minipage}\hfill
\begin{minipage}[t]{0.50\textwidth}
\vspace{0pt}
\centering
\captionsetup{type=figure}
\includegraphics[width=\linewidth]{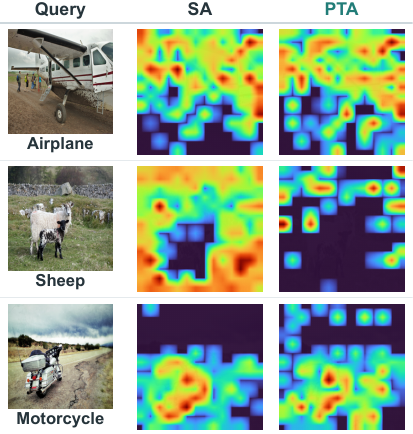}
  \caption{\textbf{Grad-CAM visualisations.} Query images and SA/PTA maps for three selected examples. Full query texts are in Appendix~\ref{app:visualisation}.}
  \label{fg:gradcam_examples}
\end{minipage}
\par

\noindent\begin{minipage}[t]{0.48\textwidth}
\vspace{0pt}
\centering
\captionsetup{type=figure}
\includegraphics[width=\linewidth]{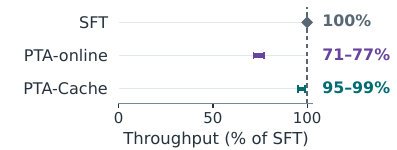}
\caption{\textbf{Training throughput.} Ranges are not confidence intervals. Online values use inverse slowdown; cache construction excluded.}
\label{fig:training-throughput}
\end{minipage}\hfill
\begin{minipage}[t]{0.48\textwidth}
\vspace{0pt}
\centering
\captionsetup{type=figure}
\includegraphics[width=\linewidth]{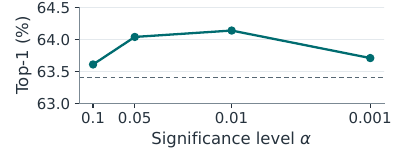}
\caption{\textbf{Sensitivity to $\alpha$.} ImageNet top-1 accuracy at the tested $\alpha$ values. Points are reported estimates; dashed: no PTA.}
\label{fig:alpha-sensitivity}
\end{minipage}
\par

\textbf{Additional experiments.} The appendix covers clean-negative ablations (Appendix~\ref{appendix:cn_ablation}); calibration budgets and $\alpha$ (Appendices~\ref{app:permutation_budget}--\ref{app:alpha_sensitivity}); low-scoring matched pairs (Appendix~\ref{app:matched_pairs_below_q}); permutation stability and held-out calibration (Appendices~\ref{app:global_permutation_stability}--\ref{app:heldout_random_pairing}); threshold rules (Appendix~\ref{app:thresholding_strategies}); teacher-quality diagnostics (Appendix~\ref{app:weak_teacher_quality}); data scaling and cached training (Appendices~\ref{app:data_scaling}--\ref{app:cache_efficiency}); and retrieval examples with full queries (Appendix~\ref{app:visualisation}).

%% file: tables/classification_results.tex
\begin{table*}[!t]
  \centering
  \small
  \definecolor{ptaTableAccent}{HTML}{1C6B65}
  \setlength{\tabcolsep}{1pt}
  \caption{\textbf{Zero-shot classification accuracy (\%).} CLIP ViT-B/16 students are fine-tuned on COCO train2014 with a CLIP ViT-B/32 teacher. Upper and lower rows show the mean and sample SD over seeds 41--45. PTA uses finite-rank calibration (implementation details). Best means are bold.}
  \label{tab:cc3m-cls}
  {\fontsize{8pt}{9.5pt}\selectfont
  \newcommand{\ptatablestd}[1]{{\scriptsize\color{black!65}$\pm$#1}}
  \newcommand{\ptatablemethod}[1]{\multirow{-2}{*}{\shortstack[l]{#1}}}
  \begin{tabular*}{\linewidth}{@{\extracolsep{\fill}}l*{13}{c}@{\hspace{4pt}}c@{}}
    \toprule
    \rowcolor[HTML]{F3F5F6}
    \textbf{Method} & \textbf{ImageNet} & \textbf{IN-V2} & \textbf{IN-A} & \textbf{IN-R} & \textbf{Cars} & \textbf{DTD} & \textbf{C-100} & \textbf{C-10} & \textbf{Food} & \textbf{CUB} & \textbf{Aircraft} & \textbf{Flowers} & \textbf{Pets} & \textcolor{ptaTableAccent}{\textbf{Avg.}} \\
    \midrule
     & 66.23 & 59.49 & 37.15 & 71.82 & 59.38 & \textbf{43.36} & 67.46 & 90.40 & 84.55 & \textbf{53.30} & 19.86 & 63.27 & 87.11 & 61.80 \\[-1pt]
    \ptatablemethod{SA} & \ptatablestd{0.05} & \ptatablestd{0.13} & \ptatablestd{0.27} & \ptatablestd{0.22} & \ptatablestd{0.23} & \ptatablestd{0.23} & \ptatablestd{0.41} & \ptatablestd{0.40} & \ptatablestd{0.13} & \ptatablestd{0.24} & \ptatablestd{0.48} & \ptatablestd{0.12} & \ptatablestd{0.17} & \ptatablestd{0.06} \\
    \addlinespace[1pt]
     & 65.78 & 58.74 & 40.89 & 72.55 & 58.29 & 43.17 & 65.88 & 89.80 & 82.58 & 49.88 & 20.15 & 62.07 & 87.63 & 61.34 \\[-1pt]
    \ptatablemethod{VPL} & \ptatablestd{0.05} & \ptatablestd{0.07} & \ptatablestd{0.31} & \ptatablestd{0.14} & \ptatablestd{0.15} & \ptatablestd{0.29} & \ptatablestd{0.51} & \ptatablestd{0.22} & \ptatablestd{0.14} & \ptatablestd{0.31} & \ptatablestd{0.17} & \ptatablestd{0.41} & \ptatablestd{0.20} & \ptatablestd{0.08} \\
    \addlinespace[1pt]
     & 65.02 & 57.76 & 40.49 & 72.15 & 56.87 & 42.19 & 65.05 & 89.63 & 81.80 & 48.05 & 18.83 & 61.39 & 87.31 & 60.50 \\[-1pt]
    \ptatablemethod{FT} & \ptatablestd{0.05} & \ptatablestd{0.06} & \ptatablestd{0.30} & \ptatablestd{0.16} & \ptatablestd{0.33} & \ptatablestd{0.27} & \ptatablestd{0.57} & \ptatablestd{0.25} & \ptatablestd{0.08} & \ptatablestd{0.22} & \ptatablestd{0.43} & \ptatablestd{0.51} & \ptatablestd{0.21} & \ptatablestd{0.06} \\
    \addlinespace[1pt]
     & 65.68 & 58.53 & 40.80 & 72.57 & 59.28 & 42.75 & 65.34 & 88.89 & 82.23 & 50.42 & 19.66 & 62.96 & 87.53 & 61.28 \\[-1pt]
    \ptatablemethod{MD} & \ptatablestd{0.10} & \ptatablestd{0.21} & \ptatablestd{0.25} & \ptatablestd{0.03} & \ptatablestd{0.23} & \ptatablestd{0.36} & \ptatablestd{0.37} & \ptatablestd{0.23} & \ptatablestd{0.08} & \ptatablestd{0.15} & \ptatablestd{0.35} & \ptatablestd{0.24} & \ptatablestd{0.17} & \ptatablestd{0.07} \\
    \addlinespace[1pt]
     & 65.77 & 58.61 & \textbf{41.28} & 72.97 & 59.28 & 42.71 & 65.36 & 89.37 & 82.38 & 50.04 & 19.58 & 62.03 & 87.65 & 61.31 \\[-1pt]
    \ptatablemethod{WSA} & \ptatablestd{0.07} & \ptatablestd{0.20} & \ptatablestd{0.28} & \ptatablestd{0.08} & \ptatablestd{0.20} & \ptatablestd{0.21} & \ptatablestd{0.49} & \ptatablestd{0.19} & \ptatablestd{0.11} & \ptatablestd{0.35} & \ptatablestd{0.47} & \ptatablestd{0.59} & \ptatablestd{0.17} & \ptatablestd{0.08} \\
    \addlinespace[1pt]
    \midrule
    \rowcolor[HTML]{E8F3F1}
     & \textbf{66.63} & \textbf{59.84} & 40.34 & \textbf{73.57} & \textbf{60.54} & 43.32 & \textbf{67.64} & \textbf{90.90} & \textbf{84.64} & 52.58 & \textbf{22.21} & \textbf{63.51} & \textbf{88.00} & \textcolor{ptaTableAccent}{\textbf{62.59}} \\[-1pt]
    \rowcolor[HTML]{E8F3F1}
    \ptatablemethod{\textcolor{ptaTableAccent}{\textbf{PTA (ours)}}} & \ptatablestd{0.07} & \ptatablestd{0.20} & \ptatablestd{0.27} & \ptatablestd{0.13} & \ptatablestd{0.27} & \ptatablestd{0.26} & \ptatablestd{0.38} & \ptatablestd{0.41} & \ptatablestd{0.10} & \ptatablestd{0.20} & \ptatablestd{0.52} & \ptatablestd{0.37} & \ptatablestd{0.28} & \ptatablestd{0.06} \\
    \bottomrule
  \end{tabular*}
  }
  \vspace{3pt}
  \begin{minipage}{\linewidth}
  \raggedright
  \fontsize{8pt}{9.5pt}\selectfont
  SA: Soft Align; VPL: VL PseudoLabel; FT: Fixed Threshold; MD: Momentum Distillation;\\
  WSA: Weak-to-Strong Augmentation. Avg. is the 13-benchmark mean; its SD uses five seed-wise averages.
  \end{minipage}
\end{table*}

%% file: tables/retrieval_results.tex
\begin{table*}[!t]
  \centering
  \small
  \definecolor{ptaTableAccent}{HTML}{1C6B65}
  \setlength{\tabcolsep}{1pt}
  \caption{\textbf{Cross-modal retrieval (recall@1/5).} All systems are fine-tuned on COCO train2014; best scores are bold. SFT: supervised fine-tuning. Other abbreviations follow Table~\ref{tab:cc3m-cls}.}
  \label{tab:cc3m-ret}
  {\fontsize{8pt}{9.5pt}\selectfont
  \begin{tabular*}{\linewidth}{@{\extracolsep{\fill}}l*{8}{c}@{}}
    \toprule
    \rowcolor[HTML]{F3F5F6}
    \cellcolor{white} & \multicolumn{4}{c}{\textbf{Flickr30k}} & \multicolumn{4}{c}{\textbf{MS-COCO}} \\
    \cmidrule(lr){2-5}\cmidrule(lr){6-9}
    \rowcolor[HTML]{F3F5F6}
    \cellcolor{white}\textbf{Method} & \multicolumn{2}{c}{\textbf{Image to Text}} & \multicolumn{2}{c}{\textbf{Text to Image}} & \multicolumn{2}{c}{\textbf{Image to Text}} & \multicolumn{2}{c}{\textbf{Text to Image}} \\
    \cmidrule(lr){2-3}\cmidrule(lr){4-5}\cmidrule(lr){6-7}\cmidrule(lr){8-9}
    \rowcolor[HTML]{F3F5F6}
    \cellcolor{white} & \textbf{R@1} & \textbf{R@5} & \textbf{R@1} & \textbf{R@5} & \textbf{R@1} & \textbf{R@5} & \textbf{R@1} & \textbf{R@5} \\
    \midrule
    SFT & 81.00 & 95.60 & 79.70 & 95.30 & 22.69 & 41.69 & 21.94 & 41.29 \\
    SA & 81.90 & 96.40 & 81.70 & 96.00 & 23.65 & 43.69 & 22.26 & 42.06 \\
    VPL & 81.30 & 96.10 & 81.50 & 95.80 & 24.91 & 45.35 & 24.24 & 44.81 \\
    FT & 83.00 & 96.20 & 81.30 & \textbf{96.40} & 23.91 & 43.72 & 23.84 & 43.69 \\
    MD & 82.70 & 96.20 & 81.20 & 96.30 & 23.89 & 43.73 & 23.69 & 43.58 \\
    WSA & \textbf{83.40} & 96.20 & 81.40 & 96.30 & 23.97 & 43.67 & 23.85 & 43.72 \\
    \midrule
    \rowcolor[HTML]{E8F3F1}
    \textcolor{ptaTableAccent}{\textbf{PTA (ours)}} & 82.40 & \textbf{96.60} & \textbf{83.20} & 96.30 & \textbf{25.24} & \textbf{45.93} & \textbf{25.06} & \textbf{45.73} \\
    \bottomrule
  \end{tabular*}
  }
  \vspace{-2pt}
\end{table*}

%% file: tables/mechanism_ablation.tex
\begin{wraptable}{r}{0.53\textwidth}
  \vspace{-8pt}
  \centering
  \small
  \caption{\textbf{PTA mechanism decomposition.} Independent ablations of Eq.~\ref{eq:total_loss}: paired accuracy gains (pp) over matched SFT controls on 13 benchmarks, mean $\pm$ SD over five seeds (three for asymmetric).}
  \label{tab:mech-ablation}
  \definecolor{ptaMechanismAccent}{HTML}{1C6B65}
  \setlength{\tabcolsep}{3pt}
  \renewcommand{\arraystretch}{1.08}
  \begin{tabular}{p{\dimexpr0.54\linewidth-2\tabcolsep\relax}>{\raggedleft\arraybackslash}p{\dimexpr0.22\linewidth-2\tabcolsep\relax}>{\raggedright\arraybackslash}p{\dimexpr0.24\linewidth-2\tabcolsep\relax}}
    \toprule
    \rowcolor[HTML]{F3F5F6}
    \textbf{Variant} & \multicolumn{2}{c}{\textbf{Avg. gain (pp)}} \\
    \midrule
    \textcolor{black!60}{SFT (ceiling)} & \multicolumn{2}{c}{\textcolor{black!60}{--}} \\
    \addlinespace[2pt]
    CN-only (App.~\ref{appendix:cn_ablation}) & $+1.076$ & $\textcolor{black!65}{\pm 0.061}$ \\
    Asymmetric (t$\to$i: CE) & $+1.147$ & $\textcolor{black!65}{\pm 0.035}$ \\
    \midrule
    \rowcolor[HTML]{E8F3F1}
    \textcolor{ptaMechanismAccent}{\textbf{PTA (full)}} & $\textcolor{ptaMechanismAccent}{\mathbf{+1.524}}$ & $\textcolor{black!65}{\pm 0.108}$ \\
    \bottomrule
  \end{tabular}
\end{wraptable}

%% file: source/appendix.tex
\section{Appendix}

\begingroup
\hypersetup{linkcolor=black}
\setlength{\parindent}{0pt}
\setlength{\parskip}{2pt}
\newcommand{\ptaAppendixContentsEntry}[2][0pt]{%
  \noindent\hspace*{#1}%
  \hyperref[#2]{\makebox[3.2em][l]{\ref*{#2}}\nameref*{#2}}%
  \nobreak\dotfill\hyperref[#2]{\pageref*{#2}}\par
}
\noindent\textbf{Contents}\par\smallskip
\ptaAppendixContentsEntry{appendix:additional_experiments}
\ptaAppendixContentsEntry[1em]{appendix:cn_ablation}
\ptaAppendixContentsEntry{app:permutation_budget}
\ptaAppendixContentsEntry{app:alpha_sensitivity}
\ptaAppendixContentsEntry{app:matched_pairs_below_q}
\ptaAppendixContentsEntry{app:global_permutation_stability}
\ptaAppendixContentsEntry{app:heldout_random_pairing}
\ptaAppendixContentsEntry{app:thresholding_strategies}
\ptaAppendixContentsEntry{app:weak_teacher_quality}
\ptaAppendixContentsEntry[1em]{app:dropout_calibration_diagnostic}
\ptaAppendixContentsEntry{app:data_scaling}
\ptaAppendixContentsEntry{app:cache_efficiency}
\ptaAppendixContentsEntry[1em]{app:compute_resources}
\ptaAppendixContentsEntry{app:visualisation}
\ptaAppendixContentsEntry[1em]{app:qualitative_retrieval}
\ptaAppendixContentsEntry[1em]{app:gradcam_queries}
\endgroup
\medskip

\subsection{Additional experiments}
\label{appendix:additional_experiments}

\subsubsection{Clean-negatives ablation}
\label{appendix:cn_ablation}
The mechanism-decomposition study in Table~\ref{tab:mech-ablation} compares the full PTA objective (Eq.~\ref{eq:total_loss}) against a \emph{clean-negatives only} variant. The CN-only variant uses the calibrated mask solely to remove false negatives from the contrastive denominator, restricting the numerator to the ground-truth diagonal:
\begin{equation}
  \ell^{\text{CN},(k)}_{\text{i2t}}
  \;=\; -\log \frac{\exp(f^{s}(I_k, T_k)/\tau)}{\exp(f^{s}(I_k, T_k)/\tau) + \sum_{m\neq k}(1-l^{\text{PTA}}_{km})\exp(f^{s}(I_k, T_m)/\tau)},
  \label{eq:cn_only}
\end{equation}
with the symmetric text-to-image counterpart. CN-only therefore isolates the contribution of \emph{not penalising} calibrated positives from the contribution of \emph{pulling} them into the numerator. Empirically, CN-only recovers $\sim$71\% of the full PTA gain in classification (Table~\ref{tab:mech-ablation}); the remaining $\sim$0.45 comes from admitting calibrated pairs into the numerator, with no additional hyperparameters.

\subsection{Ablation on the number of permutations}
\label{app:permutation_budget}
The permutation null is a Monte-Carlo estimate, so the calibrated threshold $Q$ inherits sampling variability from the size of $\mathcal{S}_{\text{perm}}$. Table~\ref{tab:perm-count-ablation} reports the across-seed variability of $Q$ as we vary the number of permuted scores from $1{,}000$ to $20{,}000$. With only $1{,}000$ scores the threshold is noticeably noisy across seeds, but at our default of $5{,}000$ scores ($R{=}5$ permutations of $1{,}000$ pairs) the across-seed range has dropped below $0.3$ in scaled-similarity units, and further increases to $10{,}000$ or $20{,}000$ produce only marginal stabilisation. We therefore use $5{,}000$ permuted scores throughout the main results; the calibration cost is negligible relative to a single training step.

\subsection{Ablation on sensitivity to \texorpdfstring{$\alpha$}{alpha}}
\label{app:alpha_sensitivity}
We study how the permutation-test significance level $\alpha$ (used to accept weakly-matched pairs as additional positives) affects downstream performance.
\figureautorefname~\ref{fig:alpha-sensitivity} reports ImageNet top-1 accuracy after fine-tuning with different $\alpha$.
Overall, performance exhibits a clear inverted-U trend: overly permissive thresholds (large $\alpha$) admit noisy positives and reduce accuracy, while overly conservative thresholds (small $\alpha$) accept too few positives and diminish the benefit of PTA.
In our setting, $\alpha=10^{-2}$ achieves the best accuracy, and $\alpha\in\{5\times10^{-2},10^{-2}\}$ remains competitive, indicating that PTA is not overly sensitive within a moderate range.
We therefore set $\alpha=10^{-2}$ as the default in all main experiments.

\begin{samepage}
\subsection{What happens to ground-truth pairs below \texorpdfstring{$Q$}{Q}?}
\label{app:matched_pairs_below_q}
A natural question is whether the calibration mistakenly excludes some true matched pairs whose weak similarity falls below the threshold. On a diagnostic over $10{,}000$ COCO training pairs evaluated with the learned threshold $Q$, around $5.6\%$ of the diagonal pairs fall below $Q$ under the weak model. The PTA selection rule preserves these pairs by construction. Specifically, each anchor's positive set always includes its original matched example; the threshold $Q$ determines which off-diagonal pairs are added as auxiliary positives. Low teacher scores therefore do not remove the original image--text pairs from supervision during student adaptation.
\par
\end{samepage}

\input{tables/permutation_count_ablation}

\begin{table}[!t]
  \centering
  \small
  \caption{Thresholding ablation. Average accuracy (\%) on a legacy benchmark suite at $B{=}4096$ with the last four vision blocks unfrozen (single seed); included to mirror prior protocol.}
  \label{tab:threshold-ablation}
  \begin{tabular}{l S[table-format=2.2] S[table-format=+1.2]}
    \toprule
    Method & {Avg. accuracy (\%)} & {$\Delta$ vs.\ SFT} \\
    \midrule
    SFT              & 62.42 &  0.00 \\
    FT               & 63.08 & +0.66 \\
    Naive percentile & 64.59 & +2.17 \\
    \textbf{PTA} & \bfseries 65.60 & \bfseries +3.18 \\
    \bottomrule
  \end{tabular}

\end{table}

\subsection{Monte Carlo stability of global permutation calibration}
\label{app:global_permutation_stability}

We fix $n=5{,}000$ COCO train2014 image--caption records and their frozen
CLIP B/32 features, using the first caption per image and fixed preprocessing.
This calibration-only diagnostic evaluates global permutations separately
from the batch-wise training implementation. For each $R\in\{1,10,100\}$,
we pool scores from $R$ uniform permutations of the complete calibration
set, discard fixed-point positions, and retain repeated scores with equal
weight. We use the linear empirical $(1-\alpha)$ quantile and the strict
decision $s>Q$. We use 30 permutation seeds
(41--70); per repetition, the $R$ settings share permutation prefixes.
The calibration features, evaluation pairs, and probe batches remain fixed.
Reference exceedance is measured on 10,000 pairs whose image and text sources
come from separate held-out record blocks. Auxiliary selection is measured
on off-diagonal pairs in ten disjoint probe batches of 512 records.
The calibration, probe, and reference record sets are disjoint.

\begin{table}[!t]
\centering
\setlength{\tabcolsep}{4pt}
\caption{Global-permutation calibration on fixed features ($n=5{,}000$).
Values are mean $\pm$ sample standard deviation across 30 permutation seeds.
$Q$ is in scaled-similarity units. Rates are percentages, with their standard deviations in percentage points.
Variability is conditional on the fixed records and features; reference
exceedance measures random-pair selection, not semantic error.}
\label{tab:global_permutation_stability}
\begin{tabular}{ccccc}
\toprule
$\alpha$ & $R$ & $Q$ &
\shortstack{Reference\\exceedance (\%)} &
\shortstack{Auxiliary\\selection (\%)} \\
\midrule
0.005 & 1   & $28.262\pm0.359$ & $0.462\pm0.067$ & $0.451\pm0.077$ \\
0.005 & 10  & $28.152\pm0.142$ & $0.473\pm0.024$ & $0.471\pm0.031$ \\
0.005 & 100 & $28.151\pm0.048$ & $0.468\pm0.008$ & $0.471\pm0.011$ \\
\midrule
0.010 & 1   & $26.520\pm0.294$ & $0.969\pm0.094$ & $0.932\pm0.106$ \\
0.010 & 10  & $26.384\pm0.126$ & $1.007\pm0.036$ & $0.977\pm0.046$ \\
0.010 & 100 & $26.387\pm0.036$ & $1.008\pm0.011$ & $0.975\pm0.013$ \\
\bottomrule
\end{tabular}
\end{table}

Table~\ref{tab:global_permutation_stability} shows that increasing $R$ reduces
conditional Monte Carlo variability. At $\alpha=0.01$, the threshold standard
deviation is approximately $8.1$ times smaller for $R=100$ than for $R=1$,
while mean reference exceedance remains near $1\%$. This experiment assesses
calibration stability, not student accuracy or semantic precision. Reusing
the same records does not create $nR$ independent calibration samples or
establish exact $\alpha$ control.

\FloatBarrier

\subsection{Empirical calibration across datasets and teachers}
\label{app:heldout_random_pairing}

We evaluate the random-pairing reference calibration on the COCO Captions
and Flickr30k training splits using pretrained CLIP ViT-B/32 and ViT-B/16
teachers in evaluation mode. We retain one caption per image. For each of
100 seeds (41--140), we randomly repartition a fixed cached feature pool into
image-disjoint calibration and held-out pools, and independently draw one
uniform caption permutation within each pool. Fixed-point positions are
removed without modifying the permutation. The calibration pool contains
20,064 image--caption records; $M$ denotes the number of retained permuted
scores used for the threshold, rather than the number of independent
observations. We take the first $M$ retained scores and set $Q$ to their
linearly interpolated empirical $(1-\alpha)$-quantile. We measure the
fraction of held-out random-pairing scores satisfying $S>Q$, using
20,000 scores per seed on COCO and 8,000 on Flickr30k.

\begin{table}[!t]
\centering
\caption{Held-out random-pairing exceedance rate (\%) for the linear-quantile
calibration rule. Entries are mean $\pm$ sample standard deviation over 100
resampling seeds; standard deviations are in percentage points. $M$ is the number of retained permuted calibration scores.
These rates concern random pairing, rather than semantic false positives.}
\label{tab:heldout_random_pairing}
\setlength{\tabcolsep}{4pt}
\begin{tabular}{cccccc}
\toprule
$\alpha$ & $M$ & \shortstack{COCO\\ViT-B/32} & \shortstack{COCO\\ViT-B/16} & \shortstack{Flickr30k\\ViT-B/32} & \shortstack{Flickr30k\\ViT-B/16} \\
\midrule
0.005 & 100 & $1.258\pm1.166$ & $1.170\pm0.869$ & $1.384\pm1.248$ & $1.176\pm0.945$ \\
0.005 & 200 & $0.909\pm0.742$ & $0.904\pm0.684$ & $0.951\pm0.817$ & $0.942\pm0.707$ \\
0.005 & 500 & $0.643\pm0.330$ & $0.631\pm0.342$ & $0.595\pm0.321$ & $0.605\pm0.309$ \\
0.005 & 1,000 & $0.560\pm0.250$ & $0.548\pm0.222$ & $0.581\pm0.293$ & $0.572\pm0.220$ \\
0.005 & 5,000 & $0.518\pm0.109$ & $0.509\pm0.101$ & $0.509\pm0.136$ & $0.525\pm0.121$ \\
0.005 & 10,000 & $0.515\pm0.089$ & $0.508\pm0.094$ & $0.496\pm0.108$ & $0.519\pm0.094$ \\
0.005 & 20,000 & $0.503\pm0.070$ & $0.503\pm0.074$ & $0.490\pm0.095$ & $0.502\pm0.084$ \\
\midrule
0.010 & 100 & $1.766\pm1.368$ & $1.675\pm1.076$ & $2.071\pm1.584$ & $1.846\pm1.285$ \\
0.010 & 200 & $1.416\pm0.907$ & $1.388\pm0.824$ & $1.480\pm1.067$ & $1.460\pm0.869$ \\
0.010 & 500 & $1.097\pm0.489$ & $1.123\pm0.478$ & $1.152\pm0.478$ & $1.146\pm0.512$ \\
0.010 & 1,000 & $1.027\pm0.326$ & $1.027\pm0.296$ & $1.087\pm0.383$ & $1.087\pm0.320$ \\
0.010 & 5,000 & $1.005\pm0.159$ & $1.006\pm0.153$ & $1.023\pm0.174$ & $1.009\pm0.173$ \\
0.010 & 10,000 & $1.009\pm0.140$ & $1.002\pm0.139$ & $1.006\pm0.140$ & $0.995\pm0.135$ \\
0.010 & 20,000 & $1.007\pm0.097$ & $1.005\pm0.110$ & $1.005\pm0.129$ & $0.985\pm0.124$ \\
\bottomrule
\end{tabular}
\end{table}

Table~\ref{tab:heldout_random_pairing} shows larger mean exceedance and
variability at small calibration budgets. With $M=20{,}000$, the mean
exceedance ranges from $0.490\%$ to $0.503\%$ at $\alpha=0.005$ and from
$0.985\%$ to $1.007\%$ at $\alpha=0.01$ across the four settings. Increasing
$M$ also reduces the observed resampling variability. These measurements
provide empirical calibration diagnostics within the tested feature pools;
they do not establish exact $\alpha$ control, conditional guarantees for an
observed calibration sample, or semantic precision. In particular, the
permuted scores share records, and the resampling repetitions reuse a
finite cached pool. This diagnostic uses a global permutation within each
pool and is separate from the minibatch calibration used in the training
experiments.

\FloatBarrier

\subsection{Ablation on thresholding strategies}
\label{app:thresholding_strategies}
PTA uses a calibrated rule to select additional in-batch positives based on weak similarities. Here we compare several thresholding strategies under an identical weak-to-strong setting, and report their average zero-shot accuracy (\tableautorefname~\ref{tab:threshold-ablation}). The variants are: (i) \emph{SFT.} Standard supervised fine-tuning on COCO without any weak-similarity filtering, (ii) \emph{FT.} A single global cosine-similarity cutoff is tuned on COCO validation and reused for all batches, (iii) \emph{Naive percentile.} The cutoff is set to the $99$th percentile of weak similarities on ground-truth COCO pairs and kept fixed during training, and (iv) \emph{PTA.} Our method derives the cutoff from the permuted null distribution of weak similarities at a chosen significance level~$\alpha$, yielding a dataset- and teacher-adaptive threshold with explicit control over the tolerated false-alarm rate. \tableautorefname~\ref{tab:threshold-ablation} summarises the seven-dataset averages for these strategies. Both FT and naive percentile improve over the SFT baseline, showing that even simple heuristics can benefit from weak similarities. However, PTA achieves the highest average accuracy among the tested thresholding strategies in this setting.

\subsection{Additional details on weak-teacher quality}
\label{app:weak_teacher_quality}
Figure~\ref{fig:weak-dropout} in the main text summarises the teacher
dropout study. Tables~\ref{tab:weak_model_zeroshot}
and~\ref{tab:weak_model_finetuning} give the per-task teacher and student
accuracies, including Caltech101~\citep{feifei2004learning} and SUN397~\citep{xiao2010sun}. We vary inference-time dropout in CLIP-B/32 over
$\{0,0.1,0.3,0.5,0.8\}$ while holding the CLIP-B/16 student architecture
and training hyperparameters fixed. The calibration-only diagnostic below
uses a separate protocol and does not attribute the student accuracy
changes to one calibration metric.

\subsubsection{Calibration diagnostics under teacher corruption}
\label{app:dropout_calibration_diagnostic}

To complement the weak-teacher accuracy study, we conduct a separate
calibration-only diagnostic with the pretrained CLIP-B/32 teacher. We vary
its inference-time dropout rate over $\{0,0.1,0.3,0.5,0.8\}$, keeping the
weights fixed. The teacher is first set to evaluation mode; modules with
dropout are then selectively activated at the requested rate. Scores are
$100$ times the cosine similarity of normalized image and text embeddings.

We fix the first $5{,}064$ eligible COCO train2014 images in image-ID order,
using the first annotated caption per image. For each of five paired seeds
($41$--$45$), we encode this fixed sample and draw one uniform global text
permutation. We discard fixed points and retain the first $M=5{,}000$
mismatched scores. This diagnostic uses the $k$-th order statistic as $Q$,
where $k=\lceil(M+1)(1-\alpha)\rceil$; the two $\alpha$ settings reuse the
same score vector. The order-statistic rule specifies the diagnostic
threshold and does not establish an exact finite-sample guarantee for
these dependent pooled scores.

We evaluate selection on a separate, fixed set of $5{,}120$ COCO val2014
images, arranged into ten seed-shuffled batches of size $B=512$. The first
caption per image defines each batch score matrix. The auxiliary selection
rate is the fraction of its $B(B-1)$ off-diagonal scores strictly above $Q$,
aggregated over the ten batches. Known-positive recall uses
separate encodings of all $25{,}610$ annotated captions for these
images to measure the fraction of corresponding image--caption scores
strictly above $Q$. This raw-score recall is distinct from the unconditional
inclusion of observed paired positives in the training mask. Across dropout
conditions, images, captions, mismatch permutations, and batch orders are
paired by seed. The fixed images are not resampled across seeds.

\begin{table}[!t]
\centering
\caption{Calibration-only weak-teacher corruption diagnostic. Entries are
mean $\pm$ sample standard deviation across five paired seeds. Auxiliary
selection and known-positive recall are percentages; their standard
deviations are in percentage points. No student is trained in this
experiment.}
\label{tab:dropout_calibration_diagnostic}
\begin{tabular}{ccccc}
\toprule
$\alpha$ & Dropout & $Q$ &
\shortstack{Auxiliary\\selection (\%)} &
\shortstack{Known-positive\\recall (\%)} \\
\midrule
0.005 & 0.0 & $27.698\pm0.623$ & $0.589\pm0.145$ & $80.72\pm5.44$ \\
0.005 & 0.1 & $27.509\pm0.537$ & $0.581\pm0.139$ & $74.49\pm4.96$ \\
0.005 & 0.3 & $26.827\pm0.376$ & $0.540\pm0.122$ & $46.70\pm5.12$ \\
0.005 & 0.5 & $27.756\pm0.157$ & $0.510\pm0.127$ & $4.88\pm0.88$ \\
0.005 & 0.8 & $25.418\pm0.178$ & $0.633\pm0.146$ & $0.65\pm0.16$ \\
\midrule
0.010 & 0.0 & $25.990\pm0.599$ & $1.143\pm0.255$ & $91.51\pm2.49$ \\
0.010 & 0.1 & $26.110\pm0.567$ & $1.056\pm0.238$ & $85.68\pm3.64$ \\
0.010 & 0.3 & $25.741\pm0.290$ & $1.059\pm0.186$ & $60.36\pm3.36$ \\
0.010 & 0.5 & $27.269\pm0.133$ & $1.041\pm0.193$ & $7.63\pm0.98$ \\
0.010 & 0.8 & $24.909\pm0.101$ & $1.119\pm0.162$ & $1.16\pm0.19$ \\
\bottomrule
\end{tabular}
\end{table}

\begin{samepage}
Table~\ref{tab:dropout_calibration_diagnostic} shows that the threshold
changes with teacher corruption, but not monotonically. At $\alpha=0.01$,
known-positive recall falls from $91.51\%$ without dropout to $1.16\%$ at
dropout $0.8$, while the off-diagonal selection rate remains $1.12\%$.
Thus, recalibration does not restore matching information lost under severe
teacher corruption. The auxiliary selection rate measures selection yield,
not the semantic precision of the selected pairs. These calibration-only
results complement the student accuracy study; they do not attribute its
accuracy changes to a single calibration metric.
\par
\end{samepage}

\pagebreak
\subsection{Ablation on data scaling}
\label{app:data_scaling}
\begin{wrapfigure}{r}{0.53\textwidth}
  \centering
  \includegraphics[width=\linewidth]{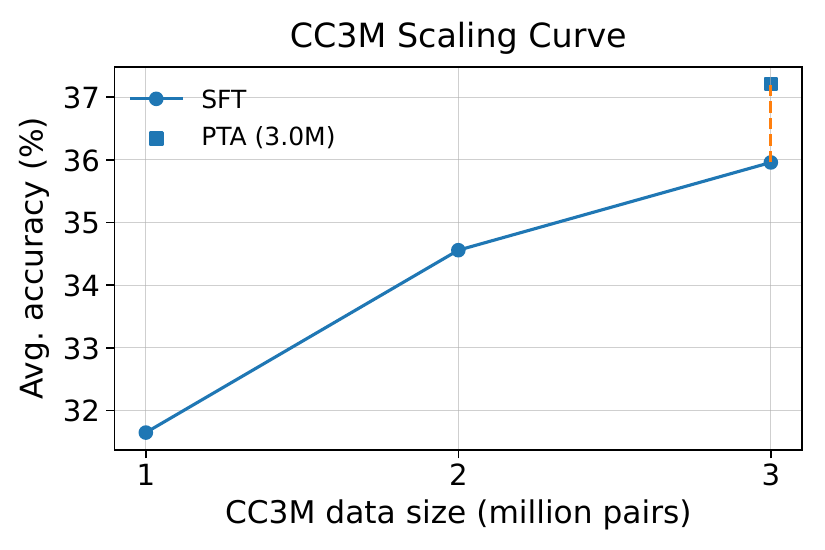}
  \caption{PTA data efficiency on CC3M.}
  \label{fig:cc3m-scaling}
\end{wrapfigure}
To evaluate whether PTA acts as a data-efficient refinement mechanism rather than merely adding computation, we conduct a scaling experiment on CC3M~\citep{sharma2018conceptual}. We conducted a pretraining on a CLIP-B/16 student for one epoch under identical hyperparameters, varying only the number of CC3M image–text pairs used: $1.0$M, $2.0$M, and $3.0$M. All baselines and PTA use the same weak teacher (CLIP-B/32) and the same training setup; the only difference is how additional positive signals are extracted. Figure~\ref{fig:cc3m-scaling} shows the average accuracy over downstream benchmarks as a function of the number of CC3M pairs. The supervised baseline exhibits clear diminishing returns: increasing the dataset from $1.0$M to $2.0$M pairs yields a $+2.91$ point gain, whereas the next million (from $2.0$M to $3.0$M) adds only $+1.40$ points. On the same $3$M examples, PTA improves performance by a further $+1.25$ points over the supervised baseline. These results highlight that PTA yields non-trivial improvements even when the training data is fixed and relatively small. Rather than requiring additional multimodal pairs, PTA provides a statistically calibrated way to reuse in-batch negatives as auxiliary positives, improving alignment in a data-efficient manner.

\subsection{PTA-Cache for efficient training}
\label{app:cache_efficiency}
To eliminate the dominant weak-model overhead, we implement a cached variant of PTA that precomputes weak embeddings once and reuses them throughout training, as the weak model is fixed and their embedding should be fixed as well.
Concretely, before fine-tuning we run the weak CLIP encoders over the entire training set to compute and store embeddings.
During training, for each mini-batch we gather the corresponding cached vectors by index, transfer them to GPU, and compute weak similarities via a single matrix multiplication; all permutation-based thresholding and positive selection is then applied directly without any additional encoder forward pass.
In practice this removes the $\sim$1.3--1.4$\times$ slowdown of PTA-online (which adds one extra CLIP forward per step) and yields per-step throughput within 1--5\% of the SFT ceiling, since the remaining costs are lightweight.

\begin{table}[!t]
  \centering
  \caption{Zero-shot classification accuracy (\%) of CLIP-B/32 with different dropout rates during inference. Higher dropout rates simulate progressively weaker models.}
  \label{tab:weak_model_zeroshot}
  \fontsize{8pt}{9.5pt}\selectfont
  \setlength{\tabcolsep}{3pt}
  \renewcommand{\arraystretch}{1.05}
    \begin{tabular}{@{}lccccccccc@{}}
      \toprule
      \makebox[51pt][l]{\textbf{Dropout}} & \textbf{Caltech101} & \textbf{CIFAR10} & \textbf{CIFAR100} & \textbf{DTD} & \textbf{Flowers102} & \textbf{Food101} & \textbf{SUN397} & \textbf{ImageNet} & \textbf{Average} \\
      \midrule
      0.0 & 59.58 & 58.38 & 42.66 & 36.52 & 60.60 & 88.71 & 80.55 & 61.00 & 61.00 \\
      0.1 & 52.77 & 50.22 & 36.06 & 39.74 & 51.66 & 89.26 & 77.36 & 56.72 & 56.72 \\
      0.3 & 28.31 & 21.74 & 26.17 & 26.36 & 38.93 & 81.08 & 50.43 & 39.00 & 39.00 \\
      0.5 & 7.32 & 4.44 & 10.96 & 12.51 & 10.02 & 33.37 & 13.09 & 13.10 & 13.10 \\
      0.8 & 0.11 & 0.50 & 2.07 & 9.77 & 1.06 & 10.32 & 0.97 & 3.54 & 3.54 \\
      \bottomrule
    \end{tabular}
\end{table}

\begin{table}[!t]
  \centering
  \caption{Classification accuracy (\%) after fine-tuning CLIP-B/16 using PTA with CLIP-B/32 weak models at different dropout rates. All use identical hyperparameters on COCO Captions.}
  \label{tab:weak_model_finetuning}
  \fontsize{8pt}{9.5pt}\selectfont
  \setlength{\tabcolsep}{3pt}
  \renewcommand{\arraystretch}{1.05}
    \begin{tabular}{@{}lccccccccc@{}}
      \toprule
      \makebox[51pt][l]{\textbf{Weak Dropout}} & \textbf{Caltech101} & \textbf{CIFAR10} & \textbf{CIFAR100} & \textbf{DTD} & \textbf{Flowers102} & \textbf{Food101} & \textbf{SUN397} & \textbf{ImageNet} & \textbf{Average} \\
      \midrule
      0.0 & 64.14 & 60.40 & 44.57 & 50.51 & 66.45 & 89.38 & 83.77 & 65.60 & 65.60 \\
      0.1 & 64.13 & 59.47 & 44.52 & 50.94 & 66.58 & 89.36 & 83.98 & 65.57 & 65.57 \\
      0.3 & 62.80 & 56.06 & 44.31 & 51.17 & 65.56 & 89.47 & 83.04 & 64.63 & 64.63 \\
      0.5 & 58.74 & 47.10 & 44.73 & 49.75 & 62.43 & 88.72 & 77.12 & 61.23 & 61.23 \\
      0.8 & 56.84 & 47.78 & 40.53 & 46.51 & 60.81 & 87.74 & 78.30 & 59.79 & 59.79 \\
      \bottomrule
    \end{tabular}
\end{table}

\subsubsection{Compute resources and recorded runtimes}
\label{app:compute_resources}
We recover hardware and elapsed times from allocation records and the
corresponding training logs and environment records.
The 30 recorded training jobs underlying Table~\ref{tab:cc3m-cls} each use
one GPU, eight CPU cores, and an 80 GB host-memory allocation. The GPU
models are NVIDIA A100 80GB PCIe, H100 80GB HBM3, and L40S 48GB; these
memory values denote nominal device capacities. The five seeds per method
follow the COCO protocol in the main text, with batch size 512, up to
20 epochs, and validation-based early stopping.

\begin{table}[!t]
  \centering
  \small
  \caption{\textbf{Recorded training allocations for Table~\ref{tab:cc3m-cls}.}
  Each job uses one GPU. Counts in parentheses indicate how many of the
  five seeds ran on each GPU model. Elapsed time is the per-job range;
  GPU-hours sum the five allocations, using unrounded times.}
  \label{tab:compute-resources}
  \setlength{\tabcolsep}{4pt}
  \begin{tabular}{@{}lrlrr@{}}
  \toprule
  Method & Runs & GPU mix across runs & Elapsed/run (h) & GPU-hours \\
  \midrule
  SA & 5 & L40S (5) & 1.08--1.35 & 5.99 \\
  VPL & 5 & L40S (5) & 1.34--1.98 & 8.15 \\
  FT & 5 & L40S (4), H100 (1) & 0.99--1.62 & 6.81 \\
  MD & 5 & L40S (5) & 1.43--1.74 & 7.99 \\
  WSA & 5 & L40S (5) & 1.44--1.67 & 7.78 \\
  \textbf{PTA} & 5 & A100 (2), H100 (3) & 1.00--2.14 & 7.24 \\
  \midrule
  \textbf{Total} & 30 & Mixed hardware & -- & \textbf{43.97} \\
  \bottomrule
  \end{tabular}
\end{table}

\textbf{Timing scope.} Job time is the allocation duration, including
setup, calibration where applicable, training, scheduled validation, and
saving; queueing and separate downstream evaluation are excluded.
GPU-hours equal allocated GPUs multiplied by elapsed hours. The 30 jobs
total 43.97 GPU-hours (4.06 A100, 4.17 H100, and 35.74 L40S). PTA's Trainer
runtime, excluding initialization and calibration but including validation,
is 53.93--87.80 minutes; its five allocations sum to 7.24 GPU-hours.
Hardware and stopping epochs differ, so these are cost records rather than
controlled speed comparisons.

\textbf{Additional cost examples.} Two separate one-epoch CC3M runs use
one A100 80GB GPU, eight CPU cores, 160 GB host memory, batch size 4096,
and seed 42, updating the full vision tower and last two text blocks;
PTA uses $\alpha=0.005$. SFT and PTA allocations are 1:38:14 and 2:06:13,
respectively (3.74 A100 GPU-hours combined). This configuration differs
from the scaling runs in Figure~\ref{fig:cc3m-scaling}.

\FloatBarrier

\subsection{Visualisation}
\label{app:visualisation}

The main text presents Grad-CAM visualisations in
Figure~\ref{fg:gradcam_examples}. Here we provide complementary retrieval
examples and the full text queries used in the visualisations.

\subsubsection{Qualitative retrieval}
\label{app:qualitative_retrieval}
\begin{figure}[!t]
  \centering
  \includegraphics[width=\linewidth]{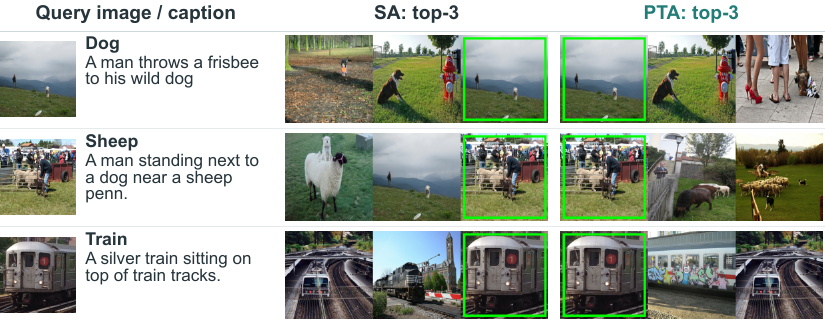}
  \caption{\textbf{Qualitative retrieval.} Three selected examples. Within
  each top-3 strip, ranks run left to right; green outlines mark the matching
  image, ranked third by SA and first by PTA.}
  \label{fg:retrieval_examples}
\end{figure}

Figure~\ref{fg:retrieval_examples} compares the top-three retrieval results
for three selected queries. Green outlines identify the matching image,
ranked third by SA and first by PTA in each example.

\subsubsection{Queries for the main-text Grad-CAM visualisations}
\label{app:gradcam_queries}
The full text queries in Figure~\ref{fg:gradcam_examples}, from top to
bottom, are:
\begin{enumerate}[leftmargin=*,itemsep=0pt,topsep=2pt]
  \item \textbf{Airplane:} A small white plane sitting on top of an airport runway.
  \item \textbf{Sheep:} Some very cute sheep in a big field.
  \item \textbf{Motorcycle:} A parked motorcycle sitting on the side of the road.
\end{enumerate}

\FloatBarrier
\clearpage

\newpage

%% file: tables/permutation_count_ablation.tex
\begin{table}[!t]
\centering
\small
\caption{\textbf{Threshold stability.} Range and standard deviation of $Q$ across seeds $\{41,42,43\}$ for different numbers of permuted scores in $\mathcal{S}_{\text{perm}}$. Threshold units are cosine similarity $\cdot\,\exp(\text{logit\_scale})$. The default is $5{,}000$ scores from $R{=}5$ random permutations of $1{,}000$ pairs.}
\label{tab:perm-count-ablation}
\begin{tabular}{lcc}
\toprule
\#\,permuted scores & Across-seed range of $Q$ & Std of $Q$ \\
\midrule
1{,}000   & 8.50\,--\,9.00 & 0.25 \\
\rowcolor[HTML]{EFEFEF}
5{,}000 (default)   & 8.44\,--\,8.69 & 0.13 \\
10{,}000  & 8.49\,--\,8.62 & 0.07 \\
20{,}000  & 8.51\,--\,8.57 & 0.03 \\
\bottomrule
\end{tabular}
\vskip -0.05in
\end{table}

%% file: checklist.tex
\section*{NeurIPS Paper Checklist}

\begin{enumerate}

  \item {\bf Claims}
  \item[] Question: Do the main claims made in the abstract and introduction accurately reflect the paper's contributions and scope?
  \item[] Answer: \answerYes{} %
  \item[] Justification: The abstract and introduction accurately reflect the contributions.

  \item {\bf Limitations}
  \item[] Question: Does the paper discuss the limitations of the work performed by the authors?
  \item[] Answer: \answerYes{} %
  \item[] Justification: Limitation is discussed in the paper.

  \item {\bf Theory assumptions and proofs}
  \item[] Question: For each theoretical result, does the paper provide the full set of assumptions and a complete (and correct) proof?
  \item[] Answer: \answerYes{} %
  \item[] Justification: Yes, we have included it.

  \item {\bf Experimental result reproducibility}
  \item[] Question: Does the paper fully disclose all the information needed to reproduce the main experimental results of the paper to the extent that it affects the main claims and/or conclusions of the paper (regardless of whether the code and data are provided or not)?
  \item[] Answer: \answerYes{} %
  \item[] Justification:We have included detailed implementation details.

  \item {\bf Open access to data and code}
  \item[] Question: Does the paper provide open access to the data and code, with sufficient instructions to faithfully reproduce the main experimental results, as described in supplemental material?
  \item[] Answer: \answerYes{} %
  \item[] Justification: We will release code upon publication.

  \item {\bf Experimental setting/details}
  \item[] Question: Does the paper specify all the training and test details (e.g., data splits, hyperparameters, how they were chosen, type of optimizer) necessary to understand the results?
  \item[] Answer: \answerYes{} %
  \item[] Justification: We have included it.

  \item {\bf Experiment statistical significance}
  \item[] Question: Does the paper report error bars suitably and correctly defined or other appropriate information about the statistical significance of the experiments?
  \item[] Answer: \answerYes{} %
  \item[] Justification: It has been reported.

  \item {\bf Experiments compute resources}
  \item[] Question: For each experiment, does the paper provide sufficient information on the computer resources (type of compute workers, memory, time of execution) needed to reproduce the experiments?
  \item[] Answer: \answerYes{}
  \item[] Justification: Appendix~\ref{app:compute_resources} reports verified hardware, GPU and host-memory allocations, and recorded times for the 30 training jobs underlying Table~\ref{tab:cc3m-cls}, plus two historical CC3M cost examples.

  \item {\bf Code of ethics}
  \item[] Question: Does the research conducted in the paper conform, in every respect, with the NeurIPS Code of Ethics \url{https://neurips.cc/public/EthicsGuidelines}?
  \item[] Answer: \answerYes{} %
  \item[] Justification: We reviewed the NeurIPS Code of Ethics and confirm this research complies.

  \item {\bf Broader impacts}
  \item[] Question: Does the paper discuss both potential positive societal impacts and negative societal impacts of the work performed?
  \item[] Answer: \answerYes{} %
  \item[] Justification: By providing a computationally efficient and statistically robust mechanism for weak-to-strong alignment, our framework lowers resource barriers to training advanced multimodal systems, though care is needed to ensure promoting uncurated auxiliary pairings does not inadvertently amplify existing societal biases present in the training data.

  \item {\bf Safeguards}
  \item[] Question: Does the paper describe safeguards that have been put in place for responsible release of data or models that have a high risk for misuse (e.g., pre-trained language models, image generators, or scraped datasets)?
  \item[] Answer: \answerNA{} %
  \item[] Justification: This work studies contrastive adaptation of existing image–text encoders using existing datasets and does not release data or models that pose a high risk for misuse.

  \item {\bf Licenses for existing assets}
  \item[] Question: Are the creators or original owners of assets (e.g., code, data, models), used in the paper, properly credited and are the license and terms of use explicitly mentioned and properly respected?
  \item[] Answer: \answerYes{} %
  \item[] Justification: Asset sources, versions, and available licenses and usage terms are documented in the \href{https://github.com/JinhaoLee/Permute-then-Adapt/blob/ef070b0cfd4a0f20f85b7033922508c9c7d11df3/THIRD_PARTY_NOTICES.md}{project repository's third-party notices}.

  \item {\bf New assets}
  \item[] Question: Are new assets introduced in the paper well documented and is the documentation provided alongside the assets?
  \item[] Answer: \answerNA{} %
  \item[] Justification: We do not introduce new datasets or models.

  \item {\bf Crowdsourcing and research with human subjects}
  \item[] Question: For crowdsourcing experiments and research with human subjects, does the paper include the full text of instructions given to participants and screenshots, if applicable, as well as details about compensation (if any)?
  \item[] Answer: \answerNA{} %
  \item[] Justification: We do not involve it.

  \item {\bf Institutional review board (IRB) approvals or equivalent for research with human subjects}
  \item[] Question: Does the paper describe potential risks incurred by study participants, whether such risks were disclosed to the subjects, and whether Institutional Review Board (IRB) approvals (or an equivalent approval/review based on the requirements of your country or institution) were obtained?
  \item[] Answer: \answerNA{} %
  \item[] Justification: We do not.

  \item {\bf Declaration of LLM usage}
  \item[] Question: Does the paper describe the usage of LLMs if it is an important, original, or non-standard component of the core methods in this research? Note that if the LLM is used only for writing, editing, or formatting purposes and does \emph{not} impact the core methodology, scientific rigor, or originality of the research, declaration is not required.
  \item[] Answer: \answerYes{} %
  \item[] Justification: Polish scripts.

\end{enumerate}